%% file: main.tex
\documentclass[12pt,twoside=false]{scrbook}
\input{header}

\begin{document}

\addchap{Detecting and Learning the Unknown in Semantic Segmentation}

\chapterauthor{
    Robin Chan\footnotemark[1],
    Svenja Uhlemeyer\footnotemark[1],
    Matthias Rottmann\footnotemark[1],
    and Hanno Gottschalk\footnotemark[1]
}

\footnotetext[1]{
    Interdisciplinary Center for Machine Learning and Data Analytics (IZMD), University of Wuppertal, Germany\\ 
    Email: \{\href{mailto:rchan@uni-wuppertal.de}{rchan}, 
    \href{mailto:suhlemeyer@uni-wuppertal.de}{suhlemeyer}, 
    \href{mailto:rottmann@uni-wuppertal.de}{rottmann}, 
    \href{mailto:hanno.gottschalk@uni-wuppertal.de}{hanno.gottschalk}\}@uni-wuppertal.de
}
    
\graphicspath{{chapters/chapter_10/figs/}}

\input{chapters/chapter_10/main}\setnewpage
\printbibliography

\end{document}

%% file: header.tex
\usepackage{listing}
\PassOptionsToPackage{export}{adjustbox}
\usepackage{amsfonts}   
\usepackage{amsmath}
\usepackage{amssymb}
\usepackage{authblk}
\usepackage{blindtext}
\usepackage{csquotes}
\usepackage{colortbl,xcolor}
\usepackage[colorlinks=true]{hyperref}
\usepackage[nameinlink,noabbrev]{cleveref}
\usepackage{ifthen}
\usepackage{verbatim}
\usepackage{xspace}
\usepackage{tocdata}
\usepackage{dsfont}
\usepackage{chngcntr}
\counterwithout{figure}{chapter}
\counterwithout{table}{chapter}
\counterwithout{equation}{chapter}
\usepackage{svg}

\usepackage[backend=biber, style=alphabetic, maxalphanames=3, backref, url=false]{biblatex}
\usepackage{graphicx}
\usepackage{pgfplots}
\pgfdeclarelayer{background}
\pgfdeclarelayer{foreground}
\pgfsetlayers{background,main,foreground} 
\pgfplotsset{compat=1.16}
\usepackage{wrapfig}
\usepackage[mode=buildnew]{standalone} 
\usepackage{tikz}
\usetikzlibrary{automata,arrows,patterns,arrows.meta,positioning,calc,shapes}
\newboolean{externalizer} 
\setboolean{externalizer}{false} 
\newcommand{\ifExtern}[2]{\ifthenelse{\boolean{externalizer}}{#1}{#2}}
\ifExtern{
\usetikzlibrary{external}
\tikzexternalize[]
}{}

\usepackage{multirow, booktabs}
\usepackage{tabu}
\usepackage{tabularx}
\usepackage{makecell}

\usepackage{algorithm} 
\usepackage{algorithmicx}
\usepackage{algpseudocode}

\newcommand{\ie}{i.e.,\ }
\newcommand{\eg}{e.g.,\ }

\newcommand{\cf}{cf.\xspace}

\newcommand{\setnewpage}{\clearpage\newpage}

\makeatletter
\renewcommand{\sectionauthor}[1]{%
  {\parindent0pt\vspace*{-5pt}%
  \linespread{1.1}\normalfont#1%
  \par\nobreak\vspace*{35pt}}
  \@afterheading%
}
\makeatother

\makeatletter
\renewcommand{\chapterauthor}[1]{%
  {\parindent0pt\vspace*{-15pt}%
  \linespread{1.1}\normalfont#1%
  \par\nobreak\vspace*{35pt}}
  \@afterheading%
}
\makeatother

\input{helpers/fewmathmacros}

\input{helpers/tu_bs_colors}

\usepackage[headsepline]{scrlayer-scrpage}
\clearpairofpagestyles

\usepackage{caption}
\usepackage[labelformat=simple, labelsep=space]{subcaption}

\newenvironment{abstract}{%
{\large\textbf{Abstract.}}\,}
{}

\automark[]{chapter}
\ihead{\normalsize\leftmark}
\ofoot{\thepage}

\makeatletter
\renewcommand{\thesection}{\@arabic\c@section}
\makeatother

\newif\ifchapone
\newif\ifchaptwo
\newif\ifchapthree
\newif\ifchapfour
\newif\ifchapfive
\newif\ifchapsix
\newif\ifchapseven
\newif\ifchapeight
\newif\ifchapnine
\newif\ifchapten
\newif\ifchapeleven
\newif\ifchaptwelve
\newif\ifchapthirteen
\newif\ifchapfourteen
\newif\ifchapfifteen
\newif\ifchapsixteen

\usepackage{layouts}

%% file: helpers/fewmathmacros.tex
\newcommand{\PRECISION}{\mathrm{Precision}} 
\newcommand{\RECALL}{\mathrm{Recall}}       
\newcommand{\mIoU}{\mathrm{mIoU}}           

\newcommand{\VEC}[1]{\mathbf{#1}}          
\newcommand{\VECG}[1]{\boldsymbol{#1}}     
\newcommand{\MAT}[1]{\mathbf{#1}}          

\newcommand{\N}{\mathbb{N}}
\newcommand{\R}{\mathbb{R}}

\newcommand{\I}{\mathbb{I}}

\newcommand{\T}[0]{\mathsf{T}}

\newcommand{\PROBD}[1][]{{\mathrm p_{#1}}} 
\newcommand{\PROB}[0]{{\mathrm P}}         
\newcommand{\EXPVAL}{\mathbb{E}}           

%% file: helpers/tu_bs_colors.tex
\definecolor{tu0}{rgb}{0.7451, 0.1176, 0.2353}

\definecolor{tu1}{rgb}{1.0000, 0.8039, 0.0000}
\definecolor{tu11}{rgb}{1.0000, 0.8627, 0.3020}
\definecolor{tu12}{rgb}{1.0000, 0.9020, 0.4980}
\definecolor{tu13}{rgb}{1.0000, 0.9412, 0.6980}
\definecolor{tu14}{rgb}{1.0000, 0.9608, 0.8000}

\definecolor{tu2}{rgb}{0.9804, 0.4314, 0.0000}
\definecolor{tu21}{rgb}{0.9882, 0.6039, 0.3020}
\definecolor{tu22}{rgb}{0.9882, 0.7137, 0.4980}
\definecolor{tu23}{rgb}{0.9922, 0.8275, 0.6980}
\definecolor{tu24}{rgb}{0.9961, 0.8863, 0.8000}

\definecolor{tu3}{rgb}{0.6902, 0.0000, 0.2745}
\definecolor{tu31}{rgb}{0.7529, 0.2000, 0.4196}
\definecolor{tu32}{rgb}{0.8431, 0.4980, 0.6353}
\definecolor{tu33}{rgb}{0.9216, 0.7490, 0.8196}
\definecolor{tu34}{rgb}{0.9529, 0.8510, 0.8902}

\definecolor{tu4}{rgb}{0.4863, 0.8039, 0.9020}
\definecolor{tu41}{rgb}{0.6431, 0.8627, 0.9333}
\definecolor{tu42}{rgb}{0.7412, 0.9020, 0.9490}
\definecolor{tu43}{rgb}{0.8431, 0.9412, 0.9686}
\definecolor{tu44}{rgb}{0.8980, 0.9608, 0.9804}

\definecolor{tu5}{rgb}{0.0000, 0.5020, 0.7059}
\definecolor{tu51}{rgb}{0.3020, 0.6510, 0.7961}
\definecolor{tu52}{rgb}{0.5490, 0.7765, 0.8667}
\definecolor{tu53}{rgb}{0.7490, 0.8745, 0.9255}
\definecolor{tu54}{rgb}{0.8510, 0.9255, 0.9569}

\definecolor{tu6}{rgb}{0.0000, 0.3255, 0.4549}
\definecolor{tu61}{rgb}{0.2510, 0.4941, 0.5922}
\definecolor{tu62}{rgb}{0.5490, 0.6941, 0.7529}
\definecolor{tu63}{rgb}{0.7490, 0.8314, 0.8627}
\definecolor{tu64}{rgb}{0.8510, 0.8980, 0.9176}

\definecolor{tu7}{rgb}{0.0314, 0.0314, 0.0314}
\definecolor{tu71}{rgb}{0.3725, 0.3725, 0.3725}
\definecolor{tu72}{rgb}{0.5882, 0.5882, 0.5882}
\definecolor{tu73}{rgb}{0.7529, 0.7529, 0.7529}
\definecolor{tu74}{rgb}{0.8667, 0.8667, 0.8667}

\definecolor{tu8}{rgb}{0.7765, 0.9333, 0.0000}
\definecolor{tu81}{rgb}{0.8431, 0.9529, 0.3020}
\definecolor{tu82}{rgb}{0.8863, 0.9647, 0.4980}
\definecolor{tu83}{rgb}{0.9333, 0.9804, 0.6980}
\definecolor{tu84}{rgb}{0.9569, 0.9882, 0.8000}

\definecolor{tu9}{rgb}{0.5373, 0.6431, 0.0000}
\definecolor{tu91}{rgb}{0.6784, 0.7490, 0.3020}
\definecolor{tu92}{rgb}{0.7686, 0.8196, 0.4980}
\definecolor{tu93}{rgb}{0.8588, 0.8941, 0.6980}
\definecolor{tu94}{rgb}{0.9059, 0.9294, 0.8000}

\definecolor{tu10}{rgb}{0.0000, 0.4431, 0.3373}
\definecolor{tu101}{rgb}{0.3020, 0.6118, 0.5373}
\definecolor{tu102}{rgb}{0.5490, 0.7490, 0.7020}
\definecolor{tu103}{rgb}{0.7490, 0.8588, 0.8353}
\definecolor{tu104}{rgb}{0.8549, 0.9176, 0.9059}

\definecolor{tu110}{rgb}{0.8000, 0.0000, 0.6000}
\definecolor{tu111}{rgb}{0.8706, 0.3490, 0.7412}
\definecolor{tu112}{rgb}{0.9216, 0.6000, 0.8392}
\definecolor{tu113}{rgb}{0.9608, 0.8000, 0.9216}
\definecolor{tu114}{rgb}{0.9804, 0.8980, 0.9608}

\definecolor{tu120}{rgb}{0.4627, 0.0000, 0.4627}
\definecolor{tu121}{rgb}{0.5961, 0.2510, 0.5961}
\definecolor{tu122}{rgb}{0.7294, 0.4980, 0.7294}
\definecolor{tu123}{rgb}{0.8392, 0.6980, 0.8392}
\definecolor{tu124}{rgb}{0.9216, 0.8510, 0.9216}

\definecolor{tu130}{rgb}{0.4627, 0.0000, 0.3294}
\definecolor{tu131}{rgb}{0.6118, 0.3020, 0.5333}
\definecolor{tu132}{rgb}{0.7569, 0.5490, 0.6980}
\definecolor{tu133}{rgb}{0.8667, 0.7490, 0.8314}
\definecolor{tu134}{rgb}{0.9216, 0.8510, 0.9020}

%% file: chapters/chapter_10/main.tex
\begin{abstract}
Semantic segmentation is a crucial component for perception in automated driving. Deep neural networks (DNNs) are commonly used for this task and they are usually trained on a closed set of object classes appearing in a closed operational domain. However, this is in contrast to the open world assumption in automated driving that DNNs are deployed to. Therefore, DNNs necessarily face data that they have never encountered previously, also known as \emph{anomalies}, which are extremely safety-critical to properly cope with.

In this work, we first give an overview about anomalies from an information-theoretic perspective. Next, we review research in detecting semantically unknown objects in semantic segmentation. We demonstrate that training for high entropy responses on anomalous objects outperforms other recent methods, which is in line with our theoretical findings.
Moreover, we examine a method to assess the occurrence frequency of anomalies in order to select anomaly types to include into a model's set of semantic categories. We demonstrate that these anomalies can then be learned in an unsupervised fashion, which is particularly suitable in online applications based on deep learning. 
\end{abstract}

\section{Introduction}
Recent developments in deep learning have enabled scientists and practitioners to advance in a broad field of applications that were intractable before. To this end, deep neural networks (DNNs) are mostly employed which are usually trained in a supervised fashion with closed-world assumption. However, when those algorithms are deployed to real-world applications, \eg artificial intelligence (AI) systems used for perception in automated driving, they often operate in an open-world setting where they have to face diversity of the real world. Consequently, DNNs are likely exposed to data which is "unknown" to them and therefore possibly beyond their capabilities to process. For this reason, having methods at hand, that indicate when a DNN is operating outside of its learned domain to seek for human intervention, is of utmost importance in safety-critical applications. 

A generic term for such a task is \emph{anomaly} detection, which is generally defined as recognizing when something departs from what is regarded as normal or common.
More precisely, identifying anomalous examples during inference, \ie new examples that are "extreme" in some sense as they lie in low density regimes or even outside of the training data distribution, is commonly referred to as \emph{out-of-distribution (OoD)} or \emph{novelty} detection in deep learning terminology.
The latter is in close connection to the task of identifying anomalous examples in training data, which is contrarily known as \emph{outlier} detection; a term originating from classical statistics to determine whether observational data is polluted. Those outlined notions are often used interchangeably in deep learning literature. Throughout this work, we will stick to the general term anomaly and only specify when distinguishing is relevant.

For the purpose of anomaly detection, plenty of methods, ranging from classical statistical ones have been developed in the past. For the nowadays most challenging computer vision tasks tackled by deep learning, where both the model weights and output are of high dimension (in the millions), specific approaches to anomaly detection are mandatory. 

Classical methods such as density estimation fail due to the curse of dimensionality. Early approaches identify outliers based on the distance to their neighbors \cite{Knorr2000,Ramaswamy2000}, \ie they are looking for sparse neighborhoods. Other methods consider relative densities to handle clusters of different densities, \eg by comparing one instance either to its $k$-nearest neighbors \cite{Breunig2000} or using an $\varepsilon$-neighborhood as reference set \cite{Papadimitriou2003}. However, the concept of neighborhoods becomes meaningless in high dimensions \cite{Aggarwal2001a}. More advanced approaches for high-dimensional data compute outlier degrees based on angles instead of distances \cite{Kriegel2008} or even identify lower-dimensional subspaces \cite{Aggarwal2001,Kriegel2009}. 

In deep-learning-driven computer vision applications, novelties are typically regarded as more relevant than outliers.
In semantic segmentation, \ie pixel-level image classification, novelty detection may even refer to a number of sub-tasks. On the one hand, we might be concerned with the detection of semantically anomalous objects. This is also known as anomaly segmentation in the case of semantic segmentation. On the other hand, we also might be concerned with the detection of changed environmental conditions that are novel. The latter may be effects of a domain shift and include change in weather, time of day, seasonality, location and time. In this work, we focus only on semantically novel objects as anomalies.

In general, an important capability of AI systems is to identify the unknown. However, when striving for improved self-reflection capabilities, anomaly detection is not sufficient. Another important capability for real-world deployment of AI systems is to realize that some specific concept appears over and over again and potentially constitutes a new (or novel) object class. Incremental learning refers to the task of learning new classes, but, especially in semantic segmentation, mostly in a strictly supervised or semi-supervised fashion where data for the new class is labeled with ground truth \cite{Michieli2019,Cermelli2020}. This is accompanied by an enormous data collection and annotation effort. In contrast to supervised incremental learning, humans may recognize a novelty of a given class that appears over and over again very well, such that in the end a single feedback might be sufficient to assign a name to a novel class. Ideally, this is accomplished in an unsupervised manner, like \cite{he2021unsupervised} for image classification.

In this work, we first introduce anomaly detection from an information-based perspective in \Cref{sec: outlier-statistics}. We provide theoretical evidence that the entropy is a suitable quantity for anomaly detection, particularly in semantic segmentation. In \Cref{sec:related-works}, we review recent developments in the fields of anomaly detection and unsupervised learning of new classes. We give an overview on existing methods, both in the context of image classification and semantic segmentation. In this setting, we examine an approach to train semantic segmentation DNNs for high entropy on anomaly data in \Cref{sec:anomaly-segmentation}. We compare this approach against other established and recent state-of-the-art anomaly segmentation methods, and empirically show the effectiveness of entropy maximization in identifying unknown objects. Lastly, we investigate an unsupervised learning technique for novel object classes in \Cref{sec: incremental-learning}. In this light, we further provide an outlook how the latter approach can be combined with entropy maximization to handle the unknown at run time in automated driving.

\section{Anomaly Detection Using Information and Entropy} \label{sec: outlier-statistics}

Anomaly detection is a common routine in any data analysis task. Before training a statistical model on data, the data should be investigated whether the underlying distribution generating the data is polluted by anomalies. In this context, anomalies can generally be understood as samples that do not fit into a distribution. Such anomalous samples can, \eg be generated in the data recording process either by extreme observations, by errors in recording and transmission, or by the fusion of datasets that use different systems of units. Most common for the detection of anomalies in statistics is the inspection of maximum and minimum values for each feature, or simple univariate visualization via box-whisker plots or histograms.

More sophisticated techniques are applied in multivariate anomaly detection. Here, anomalous samples do not necessarily have to contain extreme values for single features, but rather an untypical combination of them. One of the application areas for multivariate anomaly detection is, \eg fraud detection.   

In both outlined cases, an anomaly $\VEC{z}\in \R^d$
can be qualified as an observation that occurs at a location of extremely low density of the underlying distribution $ \PROBD(\VEC{z})$ or, equivalently, has an exceptionally high value of the information 
\begin{equation}
    I(\VEC{z})=-\log \PROBD(\VEC{z}) ~.
\end{equation} 
 
Here, two problems occur: First, it is generally not specified what is considered as exceptionally high.
Second, $\PROBD(\VEC{z})$ and thereby $I(\VEC{z})$ are generally unknown. Regarding the latter issue, however, the estimate $\hat I(\VEC{z})=-\log \hat \PROBD (\VEC{z})$ can be used which in turn relies on estimating $\hat \PROBD(\VEC{z})$ from data associated to the probability density function $\PROBD(\VEC{z})$. Estimation approaches for $\hat \PROBD(\VEC{z})$ can be distinguished between parametric and non-parametric ones.

The Mahalanobis distance \cite{mahalanobis1936generalized} is the best known parametric method for anomaly detection which is based on information of the multivariate normal distribution $N$. In fact, if $\VEC{z}\sim N(\VECG{\mu},\MAT{\Sigma})$ with mean $\VECG{\mu}\in\R^d$ and positive definite covariance matrix $\MAT{\Sigma} \in \R^{d \times d}$, then
\begin{align}
    \label{eq:Mahalanobis}
    I(\VEC{z}) &= -\log \left(\frac{1}{(2\pi)^{d/2} (\det \MAT{\Sigma})^{1/2}} \exp \Big( - \frac{1}{2} (\VEC{z}-\VECG{\mu})^\T \MAT{\Sigma}^{-1}(\VEC{z}-\VECG{\mu})\Big) \right) \\
    & = 
    \frac{d}{2}\log(2\pi) + \frac{1}{2} \log (\det \MAT{\Sigma}) + \frac{1}{2}(\VEC{z}-\VECG{\mu})^\T \MAT{\Sigma}^{-1}(\VEC{z}-\VECG{\mu}) 
    =\frac{1}{2}(\mathsf{ d}_\MAT{\Sigma}(\VEC{z},\VECG{\mu}))^2+c ~,
\end{align}
where
\begin{equation}
    \mathsf{d}_\MAT{\Sigma}(\VEC{z},\VECG{\mu}) := \sqrt{ (\VEC{z}-\VECG{\mu})^\T \MAT{\Sigma}^{-1}(\VEC{z}-\VECG{\mu}) }
\end{equation}
denotes the Mahalanobis distance.
The estimation $\hat I(\VEC{z})$ is obtained by replacing $\VECG{\mu}$ and $\MAT{\Sigma}$ by the arithmetic mean $\hat{\VECG{\mu}}$ and the empirical covariance matrix $\hat{\MAT{\Sigma}}$, respectively, and likewise $\mathsf{d}_\MAT{\Sigma}(\VEC{z}, \VECG{\mu}) $ by the empirical Mahalanobis distance $\mathsf{d}_{\hat{\MAT{\Sigma}}}(\VEC{z},\hat{\VECG{\mu}})$.

In contrast, non-parametric techniques of anomaly detection rely on non-parametric techniques to estimate $\PROBD(\VEC{z})$.  Here, a large variety of methods from histograms, kernel estimators and many others exist \cite{Klemelae2009}. We note, however, that the non-parametric estimation of densities and information generally suffers from the curse of dimensionality. To alleviate the latter issue in anomaly detection, estimation of information is often combined with techniques of dimensionality reduction, such as, \eg principal component analysis \cite{Friedman2001} or autoencoders \cite{Sakurada2014}.  

When using non-linear dimensionality reduction with autoencoders, densities obtained in the latent space depend on the encoder and not only on the data itself. This points towards a general problem in anomaly detection. If $\PROBD(\VEC{z})$ is the density of a random quantity $\VEC{z}$ and $\VEC{z}'=\VECG{\phi}(\VEC{z})$ is an equivalent encoding of the data $\VEC{z}$ using a bijective and differentiable mapping $\VECG{\phi}:\R^d\mapsto\R^d$, the \emph{change of variables} formula \cite{rudin1987, asatryan2020gan}
\begin{equation}
    \PROBD(\VEC{z}') = \PROBD(\VEC{z}) \cdot |\det \left(\MAT{J}_{\VEC{z}'} \VEC{z} \right) | = \PROBD(\VEC{z}) \cdot |\det \left(\MAT{J}_{\VEC{z}} \VECG{\phi}^{-1} (\VEC{z}) \right) |
\end{equation}
implies that
the information of $\VEC{z}'$ is 
\begin{align}
    \label{eq:InformationTransform}
    I(\VEC{z}')=-\log\left(\PROBD(\VEC{z})\right) -\log\left(\left|\det\left(\MAT{J}_\VEC{z} \VECG{\phi}^{-1}(\VEC{z})\right)\right|\right) \; ,
\end{align}
where $\MAT{J}_\VEC{z} \VECG{\phi}^{-1}(\VEC{z})$ denotes the Jacobian matrix of the inverse function $\VECG{\phi}^{-1}$. Thus, whenever a high value of $I(\VEC{z})$ indicates an anomaly, there always exists another equivalent representation of the data $\VEC{z'}$, where the information $I(\VEC{z}')$ is low. In other words, if $\VEC{z}$ is remote from other instances $\VEC{z}_j$ of a dataset and therefore considered an anomaly, there will be a transformation $\VEC{z}'=\VECG{\phi}(\VEC{z})$ that brings $\VEC{z}'$ right into the center of the data $\VEC{z}_j'=\VECG{\phi}(\VEC{z)}_j$. In fact, via the Rosenblatt transformation \cite{Rosenblatt1952} any representation $\VEC{z}$ of the data can be expressed via a representation $\VEC{z}'=\VECG{\phi}(\VEC{z})$ where $I(\VEC{z}')$ is constant over all data points. This stresses the importance to understand, that an anomaly always refers to probability \emph{and} encoding of the data $\VEC{z}$. This is true for both the original data and its approximated lower-dimensional representation.

As a side remark, autoencoders designed from neural networks have been very successfully applied in anomaly detection. Encoder and decoder networks possess the universal approximation property \cite{cybenko1989approximation}. Furthermore, common training losses like the reconstruction error are invariant under a change of the representation on latent spaces. Therefore, additional insights seem to be required to explain the empirical success of anomaly detection with autoencoders which is, however, not the scope of this work.

Another way of looking at the issue of anomaly detection in the context of different representations of same data is an explicit choice of a reference measure. This reference measure represents to which extent, or how likely, data is contaminated by potential anomalies.
Suppose we can associate the probability density $\PROBD^\text{anom}(\VEC{z})$ to the reference measure, then we can base anomaly detection on the quotient of densities, \ie the odds $\frac{\PROBD(\VEC{z})}{\PROBD^\text{anom}(\VEC{z})}$, and apply a threshold whenever this ratio is low or, equivalently, when the relative information
\begin{equation}
    \label{eq:relInfo}
    I^\text{rel}(\VEC{z}) 
    := -\log \left(\frac{\PROBD(\VEC{z})}{\PROBD^\text{anom}(\VEC{z})} \right) 
    = I(\VEC{z})-I^\text{anom}(\VEC{z})
\end{equation}
is high. We note that the relative information is independent under changes of the representation $\VEC{z}'=\VECG{\phi}(\VEC{z})$ as the $-\log|\det(\MAT{J}_\VEC{z}\VECG{\phi}^{-1}(\VEC{z}))|$ term from \Cref{eq:InformationTransform} occurs once with positive sign in $I(\VEC{z}')$ and once with negative sign in $-I^\text{anom}(\VEC{z}')$ and therefore cancels. Thus, the choice of a reference measure and the choice of a representation for the data is largely equivalent.

In practical situations, $\PROBD^\text{anom}(\VEC{z})$ is often represented by some data
$\{\VEC{z}_i^\text{anom}\}_{i\in\mathcal{T}'}$ that are either simulated or drawn from some data source of known anomalies. A binary classifier $\hat\PROBD(\text{anom}|\VEC{z})$ can then be trained on basis of proper data $\{\VEC{z}_i\}_{i\in\mathcal{T}}$ and anomalous data $\{\VEC{z}_i^\text{anom}\}_{i\in\mathcal{T}'}$. The assumed prior probability
$\PROBD(\text{anom})$
for anomalies, \ie the degree of contamination, acts as a threshold for the estimated odds. Equivalently, the estimate of the relative information
\begin{align}
    \label{eq:APriori}
    \hat I^\text{rel}(\VEC{z}) &=-\log\left(\frac{\hat \PROBD(\VEC{z})}{\hat \PROBD^\text{anom}(\VEC{z})}\right)
    \stackrel{\textrm{Bayes' Theorem}}{=} -\log\left(\frac{\hat \PROBD(\text{non-anom}|\VEC{z})\PROBD(\VEC{z})}{\PROBD(\text{non-anom})}\cdot\frac{\PROBD(\text{anom})}{\hat \PROBD(\text{anom}|\VEC{z})\PROBD(\VEC{z})}\right) \\
    &= -\log\left(\frac{1-\hat \PROBD(\text{anom}|\VEC{z})}{\hat \PROBD(\text{anom}|\VEC{z})} \cdot \frac{\PROBD(\text{anom})}{1-\PROBD(\text{anom})}\right) \\
    &=-\log\left(\frac{1-\hat \PROBD(\text{anom}|\VEC{z})}{\hat \PROBD(\text{anom}|\VEC{z})}\right) - \log\left(\frac{\PROBD(\text{anom})}{1-\PROBD(\text{anom})}\right) \\
    &=-\log\left(\frac{1-\hat \PROBD(\text{anom}|\VEC{z})}{\hat \PROBD(\text{anom}|\VEC{z})}\right) + c
\end{align}
with the prior $\log$-odds 
$c=-\log\left(\frac{\PROBD(\text{anom})}{1-\PROBD(\text{anom})}\right)$ being a parameter controlling the threshold for the binary classifier $\hat \PROBD(\text{anom}|\VEC{z})$. 

If specifying what is an exceptionally high value for the information $I(\VEC{z})$ or relative information $I^\text{rel}(\VEC{z})$, the distinction between the detection of outliers in the training data and the detection of novelties during inference has to be taken into account. In outlier detection, observations, which have high (relative) information but which are in agreement with the extreme value of the (relative) information 
\begin{equation}
\label{eq:extremal}
I^\text{max}=\max_{i\in\mathcal{T}}I(\VEC{z}_i)~~\text{or}~~I^\text{rel max}=\max_{i\in\mathcal{T}}I^\text{rel}(\VEC{z}_i),
\end{equation}
are usually intentionally not eliminated.
An outlier $\VEC{z}$ for the level of significance $0<\alpha<1$ can then be detected using the condition
\begin{equation}
    \label{eq:Outlier}
    \PROB_{\{\VEC{z}_i\}_{i\in\mathcal{T}}}(I^\text{max}>I(\VEC{z}))\leq \alpha ~~\text{or}~~\PROB_{\{\VEC{z}_i\}_{i\in\mathcal{T}}}(I^\text{rel max}>I^\text{rel}(\VEC{z}))\leq \alpha. 
\end{equation}
Note again that the distribution of $I^\text{rel}(\VEC{z}_j)$ has to be estimated to derive the associated distribution for the extreme values, see, \eg \cite{DeHaan2007}, and also $I^\text{rel}(\VEC{z})$ requires the estimation $\hat \PROBD(\VEC{z})$ or $\hat \PROBD^\text{anom}(\VEC{z})$. Therefore, a quantification of the epistemic uncertainty is essential for a proper outlier detection. Given the already mentioned problems of density estimation in high dimension, epistemic uncertainties may play a major role, unless a massive amount of data is available.

For the case of novelty detection taking place at inference, a comparison of the information of a single instance $I^\text{rel}(\VEC{z})$ with the usual distribution of information $\PROB_{\VEC{z}_i}$ seems to be in order, which leads to the novelty criterion for level of significance $0<\alpha<1$
\begin{equation}
    \label{eq:Novelty}
    \PROB_{\VEC{z}_i}(I(\VEC{z}_i)>I(\VEC{z}))\leq \alpha ~~\text{or}~~\PROB_{\VEC{z}_i}(I^\text{rel}(\VEC{z}_i)>I^\text{rel}(\VEC{z}))\leq \alpha.
\end{equation}
As a variant to this criterion, $I^\text{rel}(\VEC{z}_i)$ could also be replaced by the extreme value statistics over the number of inferences alleviating the problem of generating false novelties by multiple testing. What has been stated on the necessity to quantify the epistemic uncertainty for the case of outlier detection equally applies for novelty detection.

While anomaly detection is generally seen as a sub-field of unsupervised learning, some specific effects occur in the case of novelty detection in supervised learning. During the phase of inference, the data $\VEC{z}=(y,\VEC{x})$ contain an unobserved component $y\in\mathcal{S}$, which, \eg represent the instance's label in a classification problem for the classes contained in $ \mathcal{S}$. Using the decomposition $\PROBD(\VEC{z})=\PROBD(y,\VEC{x})=\PROBD(y|\VEC{x})\PROBD(\VEC{x})$, one obtains the (relative) information from
\begin{equation}
    \label{eq:relInformation}
    I(\VEC{z})=I(y|\VEC{x})+I(\VEC{x}),~~\text{or}~~I^\text{rel}(\VEC{z})=I(y|\VEC{x})+I^\text{rel}(\VEC{x})
    -I^\text{anom}(y|\VEC{x}),
\end{equation}
where $I(y|\VEC{x})=-\log( \PROBD(y|\VEC{x})), I^\text{anom}(y|\VEC{x})=-\log( \PROBD^\text{anom}(y|\VEC{x}))$ is the conditional information on the right hand side. Often, for the data of the reference measure $\PROBD^\text{anom}(\VEC{z})$, the labels are not contained in $\mathcal{S}$. In this case, one uses a non-informative conditional distribution $\PROBD^\text{anom}(y|\VEC{x})=\frac{1}{|\mathcal{S}|}$. If this is done, the last term of \Cref{eq:relInformation} becomes a constant that can be integrated into a threshold parameter.

The (relative) information cannot be computed without knowing $y$. Therefore, the conditional expectation is used as an unbiased estimate, yielding the expected information
\begin{equation}
    \label{eq:Entropy}
    EI(\VEC{x})=\EXPVAL_{y\sim \PROBD(y|\VEC{x})}(I^\text{rel}(\VEC{z}))= E(\VEC{x})+I^\text{rel}(\VEC{x})+b^\text{rel}, 
\end{equation}
where $E(\VEC{x})=\sum_{y\in\mathcal{S}} \PROBD(y|\VEC{x}) I(y|\VEC{x})$ is the expected information, or entropy, of the conditional distribution $\PROBD(y|\VEC{x})$ and $b^\text{rel}$ is zero for the information and equal to $-\log(|\mathcal{S}|)$ for the relative information with non-informative conditional distribution $\PROBD^\text{anom}(y|\VEC{x})$. Note that $E(\VEC{x})$ is bounded by $\log(|\mathcal{S}|)$. Therefore, under normal circumstances, the term $I^\text{rel}(\VEC{x})$ will outweigh $E(\VEC{x})$ by far. However, in problems like semantic segmentation, each component of $\VEC{x}$ is assigned a label from $\mathcal{S}$. This implies solving $|\mathcal{I}|$ classification problems, where $\mathcal{I}$ denotes the pixel space of $\VEC{x}$, thus the maximum value for $E(\VEC{x})$ yields $|\mathcal{I}|\log(|\mathcal{S}|)$.

Therefore, the first term in \Cref{eq:Entropy} contains significant contributions, especially in situations where $|\mathcal{I}|$ is large. The second term, $I^\text{rel}(\VEC{x})$ loses importance under the hypothesis that the probability of the inputs $\VEC{x}$ does not vary greatly.
Despite this hypothesis could be supported by fair sampling strategies, it requires further critical evaluation. But at least to a significant part, the expected information as an anomaly measure with regard to instance $\VEC{x}$ is given by a dispersion measure, namely the entropy of the conditional probability. As the entropy can be well estimated using a supervised machine learning approach to estimate $\hat{\PROBD}(y|\VEC{x})$ from the data $\{\VEC{z}_j\}_{j\in\mathcal{T}}$, this part of the information is well accessible in contrast to $I^\text{rel}(\VEC{x})$, which requires density estimation in high dimension.

Lastly in this section, let us give a remark on the role of anomaly data $\{\VEC{z}_j^\text{anom}\}_{j\in \mathcal{T}'} = \{\VEC{x}_j^\text{anom}\}_{j\in \mathcal{T}'}$. If such data is available, it is desirable to train the machine learning model $\hat{\PROBD}(y|\VEC{x})$ to produce high values for $E(\VEC{x}_j^\text{anom})$ so that the tractable part of the expected information $EI(\VEC{x})$ shows good separation properties. This requirement can be inserted to the loss function, as it has been proposed in \cite{Hein2019,Hendrycks2019a} for classification. In fact, as the entropy $E(\VEC{x})$ is maximized by the uniform (non-informative) label distribution $\PROBD(y|\VEC{x}_j^\text{anom})=1/|\mathcal{S}|$, the aforementioned loss will favor this prediction on anomalous inputs $\{\VEC{x}_j^\text{anom}\}_{j\in \mathcal{T}'}$. 
In this work, in \Cref{sec:anomaly-segmentation}, we will investigate this approach for the computer vision task of semantic segmentation, after having reviewed related work based on deep learning in the following section.

\section{Related Work} \label{sec:related-works}
After the introduction to anomaly detection from a theoretical point of view, we now turn to anomaly detection in deep learning. In this section, we review research in the direction of detecting and learning unknown objects in semantic segmentation.

\subsection{Anomaly Detection in Semantic Segmentation}

An emerging body of works explores the detection of anomalous inputs on image data, where the task is more commonly referred to as \emph{anomaly} or \emph{out-of-distribution (OoD) detection}. Anomaly detection was first tackled in the context of image classification by introducing post-processing techniques applied to softmax probabilities to adjust the confidence values produced by a classification model~\cite{hendrycks2017baseline, OOD, liang2018, Hein2019, Meinke2020}. These methods have proven to successfully lower confidence scores for anomalous inputs at image-level, which is why they were also adapted to anomaly detection in semantic segmentation~\cite{Angus2019, Blum2019}, \ie to \emph{anomaly segmentation} by treating each single pixel in an image as a potential anomaly. Although those methods represent good baselines, they usually do not generalize well to segmentation, \eg due to the high prediction uncertainties at object boundaries. The latter problem can, however, be mitigated by using segment-wise prediction quality estimates \cite{Rottmann18}, an approach which has also demonstrated to indicate anomalous regions within an image \cite{Oberdiek20}.

Recent works have proposed more dedicated solutions to anomaly segmentation. Among the resulting methods, many originate from uncertainty quantification. The intuition is that anomalous regions in an image correlate with high uncertainty. In this regard, early approaches estimate uncertainty using Bayesian deep learning, treating model parameters as distributions instead of point estimates~\cite{MacKay1992, Neal2012}. Due to the computational complexity, approximations are mostly preferred in practice, which comprise, \eg Monte-Carlo dropout~\cite{gal2016dropout}, stochastic batch normalization~\cite{Atanov2019}, or an ensemble of neural networks~\cite{lakshminarayanan2017simple,Gustafsson2020}; with some of them also being extended to semantic segmentation in~\cite{Badrinarayanan17BayesianSegNet, kendall2017uncertainties, Mukhoti2019}.
Even when using approximations, Bayesian models still tend to be computationally expensive. Thus, they are not well suited to real time semantic segmentation which is required for safe automated driving. 

This is why tackling anomaly segmentation with non-Bayesian methods are more favorable from a practitioner's point of view. Some approaches therefore include tuning a previously trained model to the task of anomaly detection, by either modifying its architecture or exploiting additional data. In~\cite{devries2018}, anomaly scores are learned by adding a separate branch to the neural network. In~\cite{hendrycks2019oe, Meinke2020} the network architecture is not changed but auxiliary outlier data, which is disjoint from the actual training data, is induced into the training process to learn anomalies. The latter idea motivated several works in anomaly segmentation~\cite{Blum2019, Bevandic2019a, jourdan20, Chan2020entropy}. Nonetheless, such models have to cope with multiple tasks, hence possibly leading to a performance loss with respect to the original semantic segmentation task \cite{Vandenhende2021a}. Moreover, when including outlier datasets in the training process, it cannot be guaranteed that the chosen outlier data is a good proxy for all possible anomalies.

Another recent line of works performs anomaly segmentation via generative models that reconstruct original input images. These methods assume that reconstructed images will better preserve the appearance of known image regions than that of unknown ones. Anomalous regions are then identified by means of pixel-wise discrepancies between the original and reconstructed image. Thus, such an approach is specifically designed to anomaly segmentation and has been extensively studied in~\cite{Creusot2015, Munawar2017, Lis2019, Xia2020, Lis2020, Biase2021}. The main benefit of these approaches is that they do not require any OoD training data, allowing them to generalize to all possible anomalous objects. However, all these methods are limited by the integrated discrepancy module, \ie the module that identifies relevant differences between the original and reconstructed image. In complex scenes, such as street scene images for automated driving, this might be a challenging task due to the open world setting.

Regarding the dataset landscape, only few anomaly segmentation datasets exist. The LostAndFound dataset~\cite{Pinggera2016} is a prominent example which contains anomalous objects in various streets in Germany while sharing the same setup as Cityscapes~\cite{Cordts2016}. LostAndFound, however, considers children and bicycles as anomalies, even though they are part of the Cityscapes training set. This was filtered and refined in Fishyscapes~\cite{Blum2019}. Another anomaly segmentation dataset   accompanies the CAOS benchmark~\cite{hendrycks2020scaling}, which considers three object classes from BDD100k~\cite{Yu2019} as anomalies. Both, Fishyscapes and CAOS, try to mitigate low diversity by complementing their real images with synthetic data. 

Efforts to provide anomalies in real images have been made in \cite{Lis2019} by sourcing and annotating street scene images from the web and in~\cite{Lis2020, Singh2020} by capturing and annotating images with small objects placed on the road. Just recently, the datasets published alongside the SegmentMeIfYouCan benchmark~\cite{chan2021segmentmeifyoucan} build upon those works, particularly contributing to broad diversity of anomalous street scenes as well as objects.

\subsection{Incremental Learning in Semantic Segmentation}

Building upon the detection of anomalies, training data can be enriched in order to learn novel classes. To avoid training from scratch, several approaches tackle the task of \emph{incremental} or even \emph{continuous learning}, which can be understood as adapting to continuously evolving environments. Besides learning novel classes, incremental learning also encompasses adapting to alternative tasks or other domains. A comprehensive framework to compare these different learning scenarios is provided in \cite{Ven2019}.

When learning novel classes, the primary issue incremental learning approaches face is a loss of the original performance on previously learned classes, that is commonly known as \textit{catastrophic forgetting} \cite{McCloskey1989CatastrophicII}. To overcome this problem, a model needs to be both, "stable" and "plastic", \ie the model needs to retain its original knowledge while being able to adapt to new environments. The complexity of meeting these requirements at the same time is called the stability-plasticity-dilemma \cite{Abraham2005}. In this regard, proposed solution strategies can be separated into three categories, which are either based on architecture, regularization, or rehearsal. Most of these methods have been applied to image classification first.

Architecture strategies employ separate models for each sequential incremental learning task, combined with a selector to determine which model will be used for inference \cite{Polikar2001,Chefrour2019,agarwal2020ailearn}. However, these approaches suffer from data imbalances, consequently
standard classification algorithms tend to favor the majority class. Approaches to mitigate skewed data distributions are usually based on over- or undersampling.
Another line of works, such as \cite{Rusu2016,Roy2019a}, employ "growing" models, \ie enlarging the model capacity by increasing the number of model parameters for more complex tasks. In \cite{Aljundi2017}, the authors propose an automated approach to select the proper task-specific model at test time. More efficient approaches were introduced in \cite{Gepperth2016,Yoon2018}, that restrict the adaptation of parameters to relevant parts of the model in terms of the new task. The \texttt{Self-Net} \cite{Camp2019} is made up of an autoencoder that learns low-dimensional representations of the models belonging to previously learned tasks. By that, retaining existing knowledge via approximating the old weights instead of saving them directly is accompanied with an implicit storage compression. The incremental adaptive deep model developed in \cite{Yang2019} enables capacity scalability and sustainability by exploiting the fast convergence of shallow models at the initial stage and afterwards utilizing the power of deep representations gradually. Other procedures perform continuous learning, \eg using a random-forest \cite{Hu2019}, an incrementally growing DNN, retaining a basic backbone \cite{Sarwar2020}, or nerve pruning and synapse consolidation \cite{Peng2021}.

Regularization strategies can be further distinguished between weight regularization, \ie measuring the importance of weights, and distillation, \ie transferring a model's knowledge into another. The former identifies parameters with great impact on the original tasks that are suppressed to be updated. Elastic weight consolidation (EWC) \cite{Kirkpatrick2017} is one representative method, evaluating weight importance based on the Fisher information matrix, while the  synaptic intelligence (SI) method \cite{zenke2017continual} calculates the cumulative change of Euclidean distance after retraining the model. Both regularization methods were further enhanced, \eg by combining them \cite{Chaudhry2018,Amer2019}, or by including unlabeled data \cite{Aljundi2018}. Another idea to maintain model stability was adapted in \cite{Zeng2018,Farajtabar2019}, updating gradients based on orthogonal constraints. Bayesian neural networks are applied in \cite{Lee2018a} to approximate a Gaussian distribution of the parameters from a single to a combined task. 

Distillation is a regularization method, where the knowledge of an old model can be drawn into a new model to partly overcome catastrophic forgetting. Knowledge distillation, proposed in \cite{HintonDistillation15}, was originally invented to transfer knowledge from a complex into a simple model. The earliest approach, which applies knowledge distillation to incremental learning, is called learning without forgetting (LwF) \cite{Li2018}. A combination of knowledge distillation and EWC was proposed in \cite{Schwarz2018}. Further approaches based on distillation loss are, \eg \cite{Jung2018,Yao2019,Kim2019,Lee2019}.

Rehearsal or pseudo-rehearsal-based methods, which were already proposed in \cite{Robins1995}, mitigate catastrophic forgetting by allowing the model to review old knowledge whenever it learns new tasks. While rehearsal-based methods retain a subset of the old training data, pseudo-rehearsal strategies construct a generator during retraining, which learns to produce pseudo-data as similar to the old training data as possible. Hence, they provide the advantages of rehearsal even if the previously learned information is unavailable. Methods reusing old data are, \eg incremental classifier and representation learning (iCaRL) \cite{Rebuffi2017iCaRLIC}, which simultaneously learns classifiers and feature representation, or the method presented in \cite{Castro2018EndtoEndIL}, which proposes a representative memory. The bias correction (BiC) method \cite{Wu2019} keeps old data in a similar manner, but handles the data imbalance differently. Most pseudo-rehearsal approaches include generative adversarial networks (GANs) \cite{Odena2017,Wu2018,Mellado2019,Ostapenko2019} or a variational autoencoder (VAE) \cite{Shin2017}. The method presented in \cite{Hou2018} combines distillation and retrospective (DR), whereby baseline approaches such as LwF are outperformed by a large margin.

Only few works exist, such as \cite{Tasar2019,Klingner2020ClassIncrementalLF,Michieli2021, uhlemeyer2022towards}, that adapt incremental learning techniques to semantic segmentation. Most of them adjust knowledge distillation using only a small portion or even none of old data. One challenge of continuous learning for semantic segmentation is that images may contain unseen as well as known classes. Hence, annotations that are restricted to some task assign a great amount of pixels to a background class, exhibiting a semantic distribution shift. The authors of \cite{Cermelli2020} provide a framework that mitigates biased predictions towards this background class. In general, existing works rely on supervision for incremental learning.

\section{Anomaly Segmentation} \label{sec:anomaly-segmentation}

The task of anomaly detection in the context of semantic segmentation, \ie identifying anomalies at pixel-level, is commonly known as \emph{anomaly segmentation}.
For this task several approaches have been proposed that are either based on uncertainty quantification, generative models, or training strategies specifically tailored to anomaly detection. In this work, we will first review some of those well-established methods and, subsequently, report a performance comparison with respect to their capability of identifying anomalies. In particular, we will demonstrate empirically that entropy maximization yields great performance on this segmentation task, which is in accordance to the statement of the entropy's importance from the information-based perspective as presented in \Cref{sec: outlier-statistics}.

\subsection{Methods} \label{sec:methods}

Let $\VEC{x} \in \I^{H \times W \times 3}, \I = [0,1]$, denote (normalized) color images of resolution $H \times W$. Feeding those images to a semantic segmentation network $\VEC{F}: \I^{H \times W \times 3} \to \R^{H \times W \times S} $, the model produces pixel-wise class scores $\VEC{y} = (y_{i,s})_{i\in \mathcal{I}, s\in\mathcal{S}} = \VEC{F}(\VEC{x}) \in \R^{H \times W \times S}$, with the set of pixel locations denoted by $\mathcal{I} = \{ 1, \ldots, H\} \times \{ 1, \ldots, W \}$ and the set of trained (hence known) classes denoted by $\mathcal{S} = \{ 1, \ldots, S \}$. The corresponding predicted segmentation mask is given by $\VEC{m} = (m_i)_{i \in \mathcal{I}} \in \{ 1,\ldots, S \}^{H \times W}$, where for $ m_i = \arg \max_{s\in \mathcal{S}} y_{i,s} ~ \forall ~ i\in\mathcal{I}$ the maximum a-posteriori probability principle is applied. Regarding the task of anomaly segmentation, the ultimate goal is then to obtain a score map $\VEC{a} = (a_i)_{i\in\mathcal{I}} \in \R^{H \times W}$ that indicates the presence of an anomaly at each pixel location $i \in \mathcal{I}$ within image $\VEC{x}$, \ie the higher the score the more likely there should be an anomaly.

Each of the methods employed in this section provides such score maps. Their underlying segmentation networks (\texttt{DeepLabV3+}, \cite{10.1007/978-3-030-01234-2_49}) are all trained on Cityscapes \cite{Cordts2016}, \ie objects not included in the set of Cityscapes object classes are considered as anomalies since they have not been seen during training and thus are unknown. The anomaly detection methods, however, differ in the way how the scores are obtained, which is why we briefly introduce the different techniques in the following.

\paragraph{Maximum softmax probability:} The most commonly-used baseline for anomaly detection at image level is thresholding at the maximum softmax probability (MSP) \cite{hendrycks2017baseline}. Therefore, this method assumes that anomalies are attached a low confidence or, equivalently, high uncertainty. Using MSP in anomaly segmentation, the score map is computed via
\begin{equation}
    a_i = 1 - \max_{s\in\mathcal{S}} \mathbf{softmax}(\VEC{y}_{i}) \quad \forall ~ i\in\mathcal{I} ~.
\end{equation}

\paragraph{ODIN:} A simple extension to improve MSP is applying temperature scaling as well as adding perturbations, which is known as out-of-distribution detector for Neural networks (ODIN) \cite{liang2018}. In more detail, let $t \in \R_{>0}$ be a hyperparameter for temperature scaling and let $\varepsilon \in \R_{\geq0}$ be a hyperparameter for the perturbation magnitude. Then, the input $\VEC{x}$ is modified as
\begin{equation}
    \tilde{\VEC{x}} = (\tilde{x}_i)_{i \in \mathcal{I}} \quad \textrm{with} \quad \tilde{x}_i = x_i - \varepsilon\ \mathrm{sign} \left( -\frac{\partial}{\partial x_i} \log \max_{s\in\mathcal{S}}  \mathbf{softmax}\left(\frac{\VEC{y}_{i}}{t} \right)  \right)  \quad \forall ~ i\in\mathcal{I} ~,
\end{equation}
yielding prediction $\tilde{\VEC{y}} = F(\VEC{\tilde{x}})$ for which thresholding is applied at the MSP, \ie the anomaly score map is given by
\begin{equation}
    a_i = 1 - \max_{s\in\mathcal{S}} \mathbf{softmax}(\tilde{\VEC{y}}_i) \quad \forall ~ i\in\mathcal{I} ~.
\end{equation}

\paragraph{Mahalanobis distance:} This anomaly detection approach estimates how well latent features fit to those observed in the training data. Let $(L-1)$ denote the penultimate layer of a network $\VEC{F}$ with $L$ layers. In \cite{OOD} the authors have shown that training a softmax classifier fits a class-conditional Gaussian distribution for the output features $\VEC{f}_{L-1}$. Hence, under that assumption
\begin{equation}
    \PROB \left( \VEC{y}^{(L-1)}_{i} ~\Big|~ \overline{y}_{i,s} = 1 \right) = N \left( \VEC{y}^{(L-1)}_{i} ~\Big|~ \VECG{\mu}_s, \VEC{\Sigma}_s \right) \quad \forall ~ i\in\mathcal{I} ~,
\end{equation}
where $\VEC{y}^{(L-1)} = \VEC{f}_{L-1}(\VEC{x}) \in \mathbb{R}^{H \times W \times C_{L-1}}$ denotes the feature map of the penultimate layer given input $\VEC{x}$, and $\overline{\VEC{y}}$ the corresponding one-hot encoded final target. The minimal Mahalanobis distance $d_{\VEC{\Sigma}_s}(\VEC{x}, \VECG{\mu}_s)$ is then an obvious choice for an anomaly score map
\begin{equation}
    a_i = \min_{s\in\mathcal{S}} d_{\VEC{\Sigma}_s}(\VEC{x}, \VECG{\mu}_s) = \min_{s\in\mathcal{S}} ( \VEC{y}^{(L-1)}_{i} - \VECG{\mu}_s )^\T \VEC{\Sigma}_s^{-1} ( \VEC{y}^{(L-1)}_{i} - \VECG{\mu}_s ) \quad \forall ~ i\in\mathcal{I} ~,
\end{equation}
\cf \Cref{eq:Mahalanobis}. Note that the class means $\VECG{\mu}_s \in \mathbb{R}^{C_{L-1}}$ and class covariances $\VEC{\Sigma}_s \in \mathbb{R}^{C_{L-1} \times C_{L-1}}$ are generally unknown, but can be estimated by means of the training dataset. 

\paragraph{Monte-Carlo dropout:} In semantic segmentation, Monte Carlo dropout represents the most prominent technique to approximate Bayesian neural networks. According to \cite{Mukhoti2019}, (epistemic) uncertainty is measured as the mutual information which might serve as anomaly score map, \ie
\begin{align}
    a_i = - \sum_{s \in \mathcal{S}} \left( \frac{1}{R} \sum_{r \in \mathcal{R}} \PROBD_{i,s}^{(r)} \right) \log \left( \frac{1}{R} \sum_{r \in \mathcal{R}} \PROBD_{i,s}^{(r)} \right)
    - \frac{1}{R} \sum_{s \in \mathcal{S}} \sum_{r \in \mathcal{R}} \PROBD_{i,s}^{(r)} \log \PROBD_{i,s}^{(r)} \quad \forall ~ i\in\mathcal{I},
\end{align}
with $\VEC{p}_{i}^{(r)} = (p_{i,s}^{(r)})_{s\in\mathcal{S}} = \mathbf{softmax}(\VEC{y}_{i}^{(r)})$ in the sampling round $r \in \mathcal{R} = \{ 1,\ldots R\}$. Typically, $8 \leq R \leq 12$.

\paragraph{Void classifier:} Neural networks can be trained to output confidences for the presence of anomalies \cite{devries2018}. One approach in this context is adding an extra class to the set $\mathcal{S}$ of previously trained classes of a semantic segmentation network, which then also requires annotated anomaly data to learn from. To this end, the void class in Cityscapes is a popular choice as proxy for all possible anomaly data \cite{Blum2019}, in particular if the segmentation model was originally trained on Cityscapes. Thus, the softmax output of the additional class $s=S+1$ represents the anomaly score map, \ie
\begin{equation}
    a_i = \mathrm{softmax}_{s=S+1}(\VEC{y}'_{i}) \quad \forall ~ i\in\mathcal{I} ~, 
\end{equation}
where $\VEC{y}' = (\VEC{y}'_i)_{i\in \mathcal{I}} = (y'_{i,s})_{i\in \mathcal{I}, s\in\{1,\ldots, S+1\}} = \VEC{F}'(\VEC{x}), \VEC{F}': \I^{H \times W \times 3} \to \R^{H \times W \times (S+1)}$.

\paragraph{Learned embedding density:} Let $\VEC{f}_{\ell}(\VEC{x}) \in \R^{H_\ell \times W_\ell \times C_\ell}$ denote the feature map, or equivalently feature embedding, at layer $\ell \in \mathcal{L}=\{1,\ldots,L\}$ of a semantic segmentation network. By employing normalizing flows, the true distribution of features $\mathbf{p}(\VEC{f}_{\ell}(\VEC{x})) \in \I^{H_\ell \times W_\ell}$, where $\VEC{x} \in \mathcal{X}^\mathrm{train}$ is drawn from the training dataset, can be trained via maximum likelihood, \ie normalizing flows learn to produce the approximation $\hat{\mathbf{p}}(\VEC{f}_{\ell}(\VEC{x})) \approx \mathbf{p}(\VEC{f}_{\ell}(\VEC{x}))$ \cite{Blum2019}. At test time, the negative log-likelihood measures how well features of a test sample fit to the feature distribution observed in the training data, yielding the anomaly score map
\begin{equation}
     \VEC{a} = \mathbf{up}^\mathrm{lin} \left( -\VECG{\log} ~ \hat{\mathbf{p}}(\VEC{f}_{\ell}(\VEC{x})) \right) \quad \textrm{($\VECG{\log}$ applies $\log$ element-wise)}
\end{equation}
with $\mathbf{up}^\mathrm{lin}: \R^{H_\ell \times W_\ell} \to \R^{H\times W}$ denoting (bi-)linear upsampling.

\paragraph{Image resynthesis:} After obtaining the predicted segmentation mask $\VEC{m} \in \{ 1,\ldots,S\}^{H \times W}$, $\VEC{m} = (m_i)_{i\in \mathcal{I}}$, this output can be further processed by a generative model $\VEC{G}:\{ 1,\ldots,S\}^{H \times W} \to \I^{H \times W \times 3} $ aiming to reconstruct the original input image, \ie $\VEC{x}'=\VEC{G}(\VEC{m}) \approx \VEC{x}$. This process is also called image resynthesis, and the intuition is that reconstruction quality for anomalous objects is worse than for those on which the generative model is trained on. To determine pixel-wise anomalies, a discrepancy network \cite{Lis2019} $\VEC{D}:\{ 1,\ldots,S\}^{H \times W} \times \I^{H \times W \times 3} \times \I^{H \times W \times 3} \to \R^{H \times W}$ can then be employed, which classifies whether one pixel is anomalous or not, based on information provided by $\VEC{m}, \VEC{x}'$, and $\VEC{x}$. 
Here, $\VEC{D}$ is trained on intentionally triggered classification mistakes that are produced by flipping classes on predicted segmentation masks.
The anomaly score map is given by the output of the discrepancy network, \ie
\begin{equation}
    \VEC{a} = \VEC{D}(\VEC{m}, \VEC{x}', \VEC{x}) = \VEC{D}(\VEC{m}, \VEC{G}(\VEC{m}), \VEC{x}) ~.
\end{equation}

\paragraph{SynBoost:} The image resynthesis approach is limited by the employed discrepancy module $\VEC{D}$. In \cite{Biase2021}, the authors proposed to extend the discrepancy network by incorporating further inputs based on uncertainty, such as the pixel-wise softmax entropy 
\begin{equation} \label{eq:softmax-entropy}
    H_i(\VEC{x}) = -\sum_{s \in \mathcal{S}} \mathrm{softmax}_s(\VEC{y}_{i}) \log(\mathrm{softmax}_s(\VEC{y}_{i}))  \quad \forall ~ i\in\mathcal{I},
\end{equation}
and the pixel-wise softmax probability margin
\begin{equation}
    M_i(\VEC{x}) = 1 - \max_{s\in\mathcal{S}} \left( \mathbf{softmax}(\VEC{y}_{i}) \right) + \max_{s\in\mathcal{S} \setminus \{ m_i\}} \left( \mathbf{softmax}(\VEC{y}_{i}) \right) \quad \forall ~ i\in\mathcal{I} ~.
\end{equation}
Furthermore, $\VEC{D}$ is trained on anomaly data provided by the Cityscapes void class.
Thus, the anomaly score map is given by 
\begin{equation}
    \VEC{a} = \VEC{D}(\VEC{m}, \VEC{x}', \VEC{x}, \VEC{H}(\VEC{x}), \VEC{M}(\VEC{x})) = \VEC{D}(\VEC{m}, \VEC{G}(\VEC{m}), \VEC{x}, \VEC{H}(\VEC{x}), \VEC{M}(\VEC{x})) ~.
\end{equation}
with $\VEC{H}(\VEC{x}) = (H_i(\VEC{x}))_{i\in\mathcal{I}}$ and $\VEC{M}(\VEC{x}) = (M_i(\VEC{x}))_{i\in\mathcal{I}}$.

\paragraph{Entropy maximization:} \label{sec:entropy-max} A desirable property of semantic segmentation networks is that they attach high prediction uncertainty to novel objects. To this end, the softmax entropy, see \Cref{eq:softmax-entropy}, is one intuitive uncertainty measure. The segmentation network can be trained for high entropy on anomalous inputs via the multi-criteria training objective \cite{Chan2020entropy}
\begin{equation}
    J^\mathrm{total} = (1-\lambda)\, \EXPVAL_{(\overline{\VEC{y}}, \VEC{x}) \sim \mathcal{D}} \big[J^\mathrm{CE}(\VEC{F}(\VEC{x}), \overline{\VEC{y}})\big] + \lambda\, \EXPVAL_{\VEC{x} \sim \mathcal{D^\mathrm{anom}}} \big[J^\mathrm{anom}(\VEC{F}(\VEC{x}))\big]~,
\end{equation}
where $\mathcal{D}$ denotes non-anomaly training data (labels available) and $\mathcal{D^\mathrm{anom}}$ denotes anomaly training data (no labels available). In this approach, the COCO dataset \cite{Lin2014} represents a set of so-called \emph{known unknowns}, which is used as proxy for $\mathcal{D^\mathrm{anom}}$ with the aim to represent all possible anomaly data.
Moreover, $\lambda \in \I$ is a hyperparameter controlling the impact of each single loss function on the overall loss $J^\mathrm{total}$. For non-anomaly data, the loss function is chosen to be the commonly-used cross-entropy $J^\mathrm{CE}$, while for anomaly data, \ie for known unknowns, we have
\begin{equation}
    J^\mathrm{anom}(\VEC{F}(\VEC{x})) = - \frac{1}{H \cdot W} \sum_{i \in \mathcal{I}} \frac{1}{S} \sum_{s \in \mathcal{S}} \log \mathrm{softmax}_s(\VEC{y}_{i})~, \quad \VEC{x} \sim \mathcal{D^\mathrm{anom}} ~.
\end{equation}
Therefore, minimizing $J^\mathrm{anom}$ is equivalent to maximizing the softmax entropy since both reach their optimum when the softmax probabilities are uniformly distributed, \ie $\mathrm{softmax}_s(\VEC{y}_{i}) = \frac{1}{S} ~ \forall ~ s\in\mathcal{S}, i\in\mathcal{I}$. After training, the anomaly score map is then given by the (normalized) softmax entropy
\begin{equation}
    a_i = \frac{1}{\log S} H_i(\VEC{x}) = - \frac{1}{\log S}\sum_{s \in \mathcal{S}} \mathrm{softmax}_s(\VEC{y}_{i}) \log(\mathrm{softmax}_s(\VEC{y}_{i}))  \quad \forall ~ i\in\mathcal{I} ~.
\end{equation}
From an information-based point of view, the entropy contains significant contribution to the expected information. This particularly applies for instance predictions in semantic segmentation, which motivates the entropy maximization approach for the detection of unknown objects, \cf \Cref{sec: outlier-statistics}.

\subsection{Evaluation and Comparison of Anomaly Segmentation Methods}
Discriminating between anomaly and non-anomaly is essentially a binary classification problem. In order to evaluate the pixel-wise anomaly detection capability, we use the receiver operating characteristic (ROC) curve as well as the precision recall (PR) curve. While for the ROC curve the true positive rate is plotted against the false positive rate at varying thresholds, in the PR curve precision is plotted against recall at varying thresholds. 
Note that we consider anomalies as the positive class, \ie correctly identified anomaly pixels are considered as true positive.
In both curves, the degree of separability is then measured by the area under the curve (AUC), where better separability corresponds to a higher AUC.

The main difference between these two performance metrics is how they cope with class imbalance. While the ROC curve incorporates the number of true negatives (for the computation of the false positive rate), in PR curves true negatives are ignored and, consequently, more emphasis is put on finding the positive class. With the anomaly score maps as defined in \Cref{sec:methods}, in our case, finding the positive class corresponds to identifying anomalies.

\begin{figure}[!t]
    \centering
    \captionsetup[subfigure]{labelformat=empty, position=top}
    \subfloat[Input \& annotation]{\includegraphics[width=0.199\textwidth]{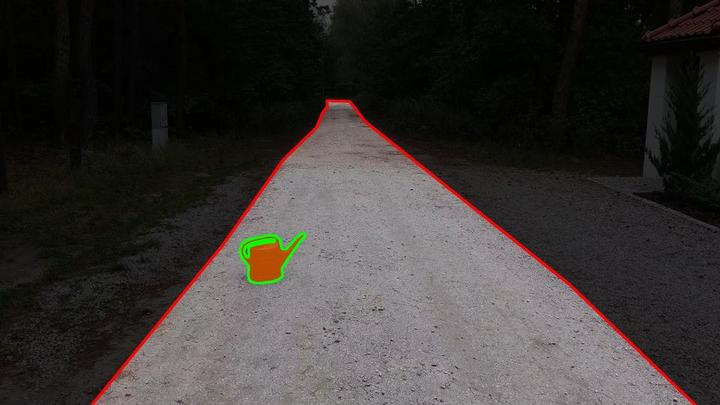}}
    \subfloat[MSP]{\includegraphics[width=0.199\textwidth]{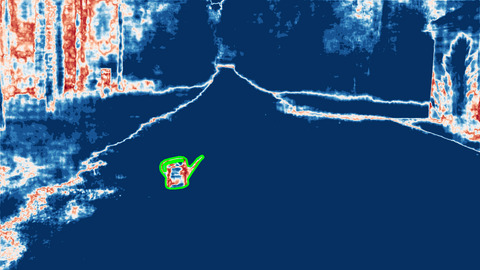}}
    \subfloat[ODIN]{\includegraphics[width=0.199\textwidth]{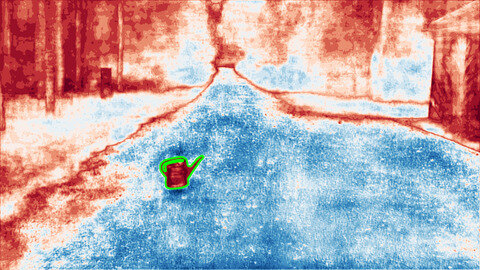}}
    \subfloat[Mahalanobis]{\includegraphics[width=0.199\textwidth]{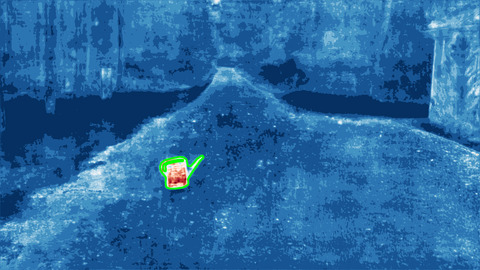}}
    \subfloat[MC dropout]{\includegraphics[width=0.199\textwidth]{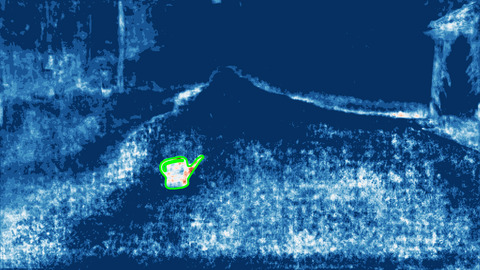}}\\
    \vspace{-.05\baselineskip}
    \captionsetup[subfigure]{labelformat=empty, position=bottom}
    \subfloat[Void classifier]{\includegraphics[width=0.199\textwidth]{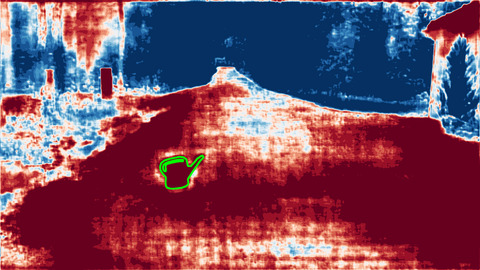}}
    \subfloat[Embedding density]{\includegraphics[width=0.199\textwidth]{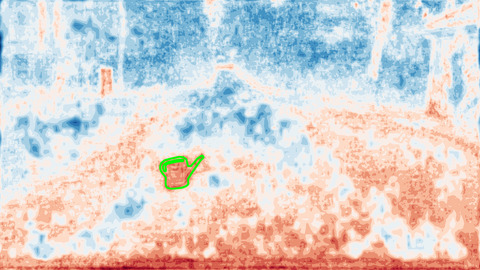}}
    \subfloat[Image resynthesis]{\includegraphics[width=0.199\textwidth]{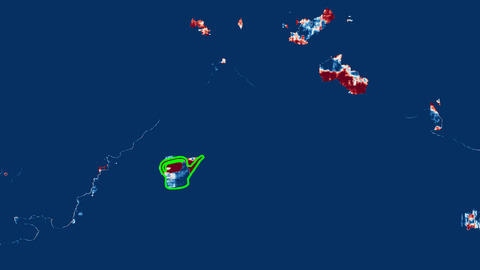}}
    \subfloat[SynBoost]{\includegraphics[width=0.199\textwidth]{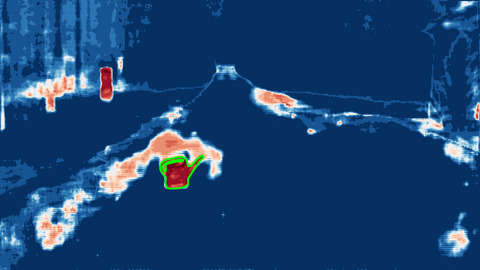}}
    \subfloat[Entropy max.]{\includegraphics[width=0.199\textwidth]{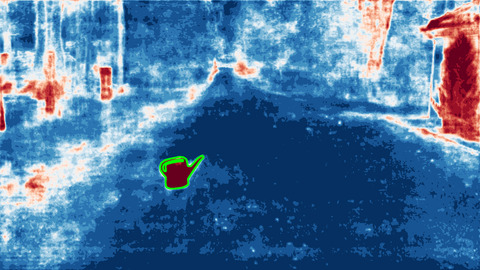}}
    \caption{Qualitative comparison of anomaly score maps for one example image of RoadAnomaly21. Here, red indicates high anomaly scores while blue indicates low ones. The ground truth anomaly instance is highlighted by green contours. Note that the region of interest is restricted to the road, highlighted by red contours in the annotation. 
    }
    \label{fig:anomaly-maps-comparison}
\end{figure}

\begin{figure}[!t]
    \centering
    ~~~~~~
    \includegraphics[width=.9\textwidth]{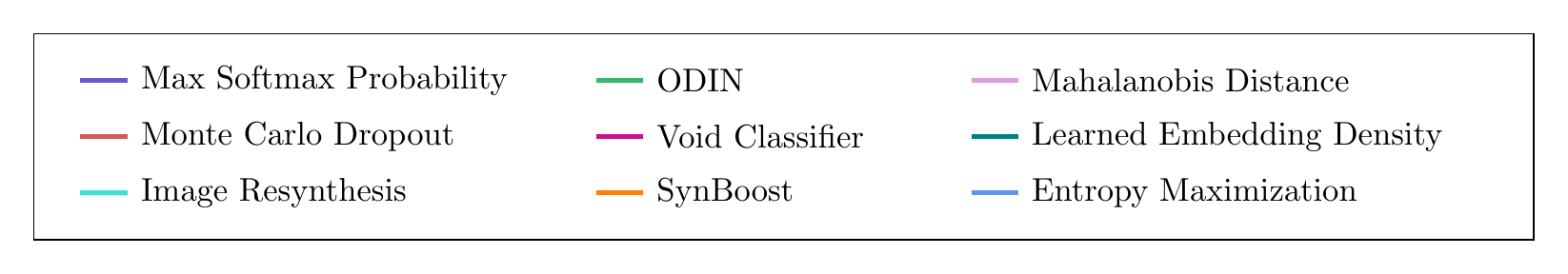}
    \vspace{.5cm}
    \includegraphics[scale=0.9]{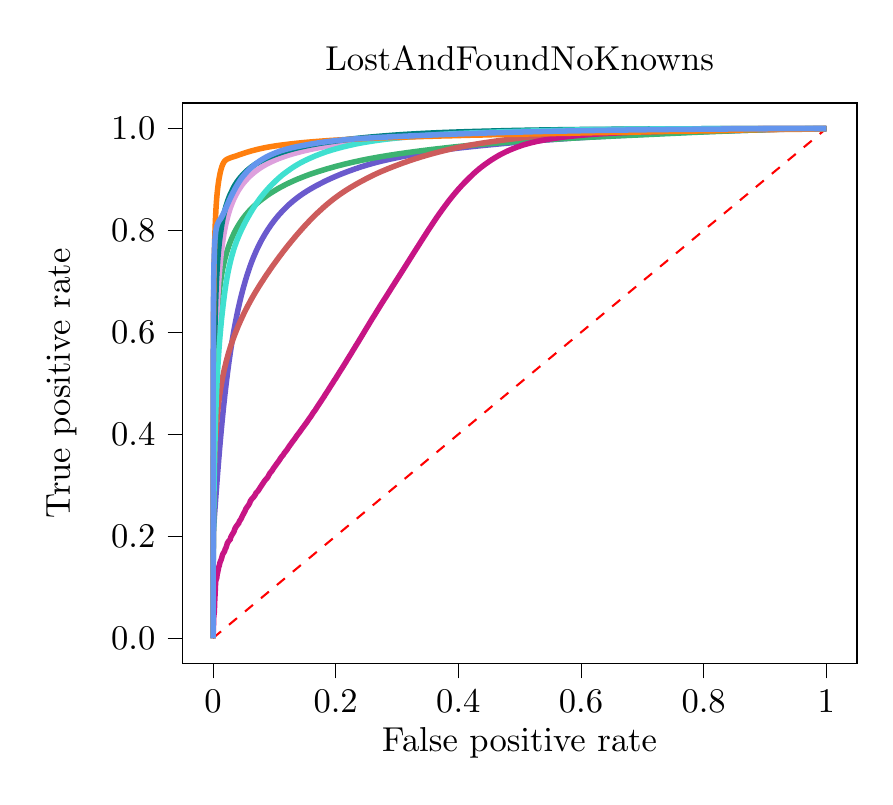}
    \includegraphics[scale=0.9]{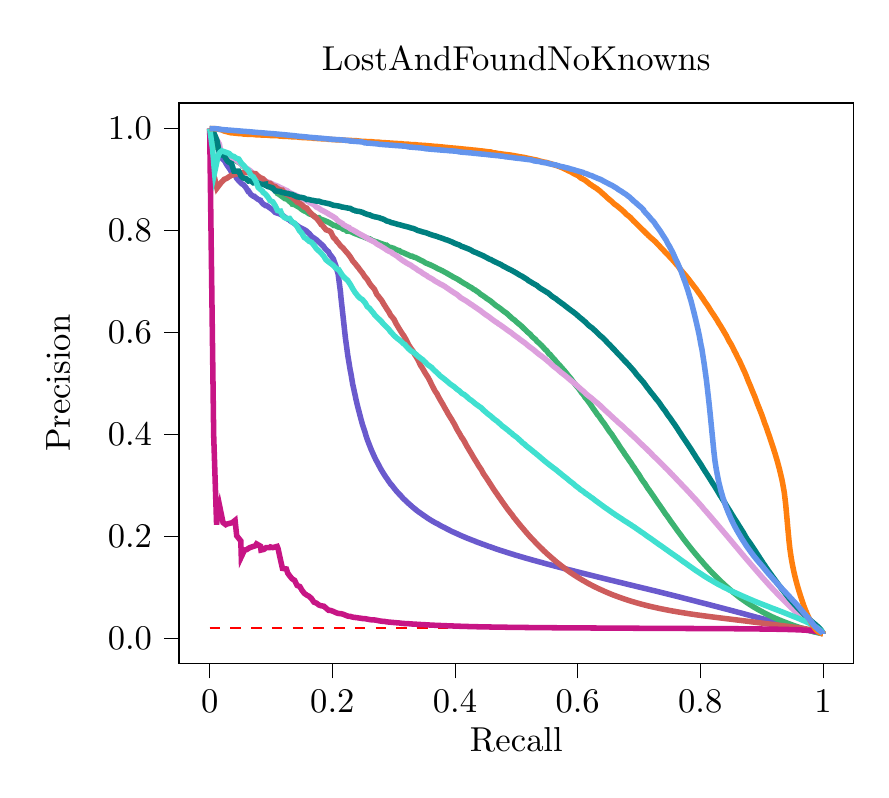} \\
    \includegraphics[scale=0.9]{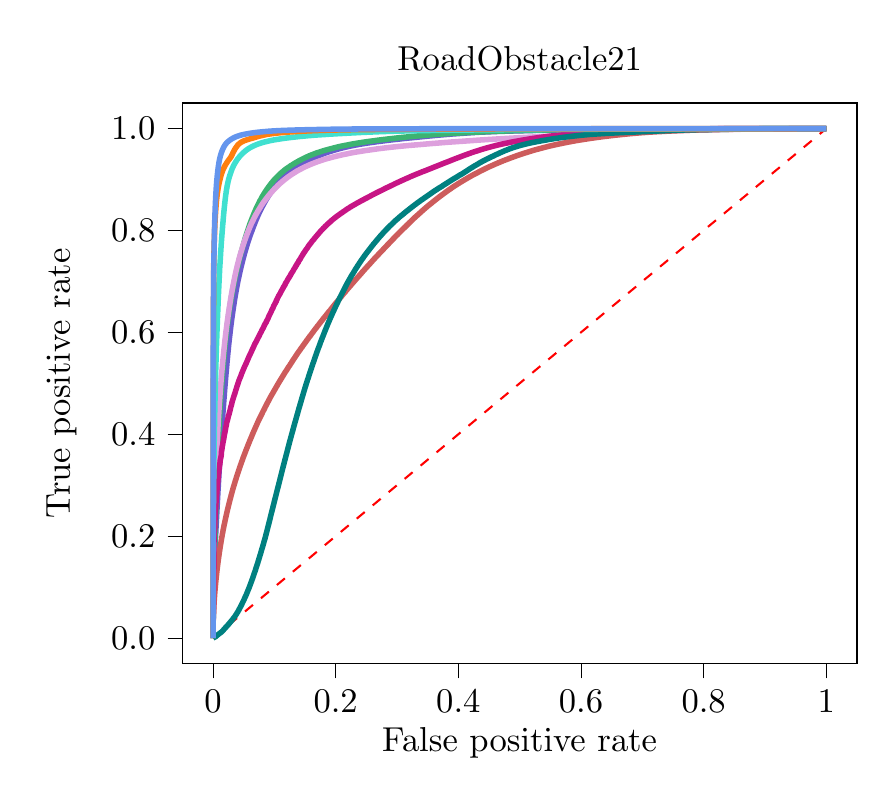}
    \includegraphics[scale=0.9]{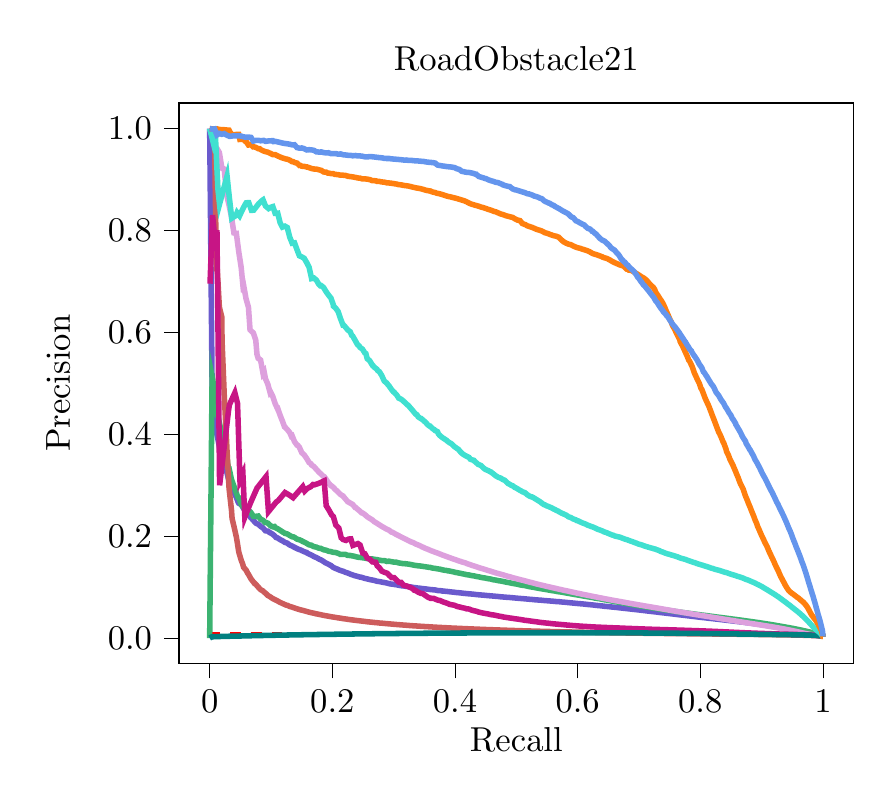}
    \caption{Receiver operating characteristic (left column) and precision recall (right column) curves for LostAndFoundNoKnowns (top row) and RoadObstacle21 (bottom row), respectively. Dashed red lines indicate the performance of random guessing, \ie the "no-skill" baseline. The degree of separability between anomaly and non-anomaly is measured by the area under the curve.}
    \label{fig:curves-anomaly-seg}
\end{figure}

As evaluation datasets, we use LostAndFoundNoKnown \cite{Blum2019} and RoadObstacle21 \cite{chan2021segmentmeifyoucan}, which are both part of the public SegmentMeIfYouCan\footnote{\url{www.segmentmeifyoucan.com}} anomaly segmentation benchmark. LostAndFoundNoKnown consists of 1043 road scene images where obstacles are placed on the road. This dataset is a subset of the prominent LostAndFound dataset \cite{Pinggera2016} but considers only obstacles from object classes which are disjoint to those in the Cityscapes labels \cite{Cordts2016}. More precisely, images with humans and bicycles are removed such that the remaining obstacles in the dataset also represent anomalies to models trained on Cityscapes. Similar scenes can be found in RoadObstacle21. That dataset was published alongside the SegmentMeIfYouCan benchmark and contains 327 road obstacle scene images with diverse road surfaces as well as diverse types of anomalous objects. Both datasets restrict the region of interest to the road where anomalies appear. This task is extremely safety-critical as it is mandatory in automated driving to make sure that the drivable area is free of any hazard. All anomaly segmentation methods introduced in the preceding \Cref{sec:methods} are suited to be evaluated on these datasets. We provide a visual comparison of anomaly scores produced by the tested methods in \Cref{fig:anomaly-maps-comparison}. We report numerical results in \Cref{tab:curves-anomaly-seg} and in the corresponding \Cref{fig:curves-anomaly-seg}. \vfill

\begin{table}[!t]
\centering
\scalebox{0.9}{
\begin{tabular}{l|ccc|ccc}
\hline
{} & \multicolumn{3}{c|}{\textbf{LostAndFoundNoKnown}} & \multicolumn{3}{c}{\textbf{RoadObstacle21}} \\
Method &  AuPRC $\uparrow$ &  AuROC $\uparrow$ &  FPR$_{95\text{TPR}}$ $\downarrow$ &  AuPRC $\uparrow$ &  AuROC $\uparrow$ &  FPR$_{95\text{TPR}}$ $\downarrow$ \\
\hline
\hline
Maximum Softmax         & 30.1  & 93.0  & 33.2  & 10.0  & 95.5  & 17.9 \\
ODIN                    & 52.9  & 95.1  & 30.0  & 11.9  & 96.0  & 16.4 \\
Mahalanobis             & 55.0  & 97.5  & 12.9  & 19.5  & 95.1  & 21.7 \\
Monte Carlo Dropout     & 36.8  & 92.2  & 35.5  & 4.9   & 83.5  & 50.3 \\
Void classifier         & 4.8   & 79.5  & 47.0  & 10.4  & 89.7  & 41.5 \\
Embedding density       & 61.7  & 98.0  & 10.4  & 0.8   & 81.0  & 46.4 \\
Image resynthesis       & 42.7  & 96.4  & 17.4  & 37.5  & 98.6  & 4.7 \\
SynBoost                & \textbf{81.7}  & \textbf{98.3}  & \textbf{4.6}   & 71.3  & 99.4  & 3.2 \\
Entropy maximization    & 77.9  & 98.0  & 9.7   & \textbf{76.0}  & \textbf{99.7}  & \textbf{1.3} \\
\hline
\end{tabular}
}
\caption{Pixel-wise anomaly detection performance on the datasets LostAndFoundNoKnown and RoadObstacle21, respectively. The main evaluation metric represents the area under precision-recall curve (AuPRC). Moreover, the area under receiver operating characteristic (AuROC) and the false positive rate at a true positive rate of 95\% (FPR$_{95\text{TPR}}$) are reported for further insights.} \label{tab:curves-anomaly-seg}
\end{table}

In general, we observe that anomaly detection methods originally designed for image classification, including MSP, ODIN and Mahalanobis, do not generalize well to anomaly segmentation. As the Mahalanobis distance is based on statistics of the Cityscapes dataset, the anomaly detection is likely to suffer from performance loss under domain shift. The same holds for Monte Carlo dropout and learned embedding density, particularly resulting in poor performance in RoadObstacle21, where various road surfaces are available. Therefore, those methods potentially act as domain shift classifier rather than as detector of unknown objects. 

The detection methods based on autoencoders, namely image resynthesis and SynBoost, show to be better suited for the task of anomaly segmentation, clearly being superior to all the approaches that already have been discussed. Autoencoders are limited by their discrepancy module, and we observe that anomaly detection performance significantly benefits from incorporating uncertainty measures, as done by SynBoost. Only entropy maximization reaches similar anomaly segmentation performance, even outperforming SynBoost in RoadObstacle21. This again can be explained by the diversity of road surfaces, which detrimentally affects the discrepancy module. 

As a final remark, we draw attention to the use of anomaly data. The void classifier follows the same intuition as entropy maximization by including known unknowns, but cannot reach nearly as good anomaly segmentation performance. We conclude that the COCO dataset is better suited as proxy for anomalous objects than the Cityscapes unlabeled objects. Moreover, the results of that method empirically demonstrate the impact of the entropy in anomaly segmentation, which is in accordance to the statement of the entropy's importance from the information perceptive described in \Cref{sec: outlier-statistics}.

\begin{table*}
    \begin{center}
    \scalebox{0.9}{
    \begin{tabular}{c|rrrr|rrrr}
    \hline
    Anomaly score / entropy & \multicolumn{4}{|c}{Entropy maximization} & \multicolumn{4}{|c}{Entropy maximization} \\
    threshold & \multicolumn{4}{|c}{+ thresholding}  & \multicolumn{4}{|c}{+ thresholding + meta classifier}  \\
    $a_i \geq \tau, i\in\mathcal{I}$ & FP $\downarrow$ & FN $\downarrow$  & F$_1$ $\uparrow$ & $\delta$ in \% $\downarrow$ & FP $\downarrow$ & FN $\downarrow$  & F$_1$ $\uparrow$ & $\delta$ in \% $\downarrow$ \\
    \hline
    \hline
    $\tau=0.30$    & 8,068  & 191   & 0.26  & 0.30  & 290  & 308 & 0.82 & 0.06 \\
    $\tau=0.40$    & 4,035  & 289   & 0.39  & 0.11  & 251  & 359 & 0.81 & 0.03 \\
    $\tau=0.50$    & 1,215  & 415   & 0.60  & 0.04  & 145  & 447 & 0.80 & 0.02 \\
    $\tau=0.60$    & 327    & 613   & 0.69  & 0.02  & 49   & 619 & 0.76 & 0.02 \\
    $\tau=0.70$    & 135    & 879   & 0.61  & 0.01  & 21   & 881 & 0.63 & 0.01 \\
    \hline
    \end{tabular}
    }
    \end{center}
    \caption{Detection errors at object level for LostAndFound anomalies at different anomaly score / entropy thresholds $\tau$ (to generate anomaly segmentation masks). The quantities false-positives (FP) and false-negatives (FN) are reported at segment level, with anomalies as positive class. The F$_1$ summarizes these quantities into an overall measure. By $\delta$ we denote the performance loss on the original task, which is the semantic segmentation of Cityscapes. In this context, we consider a performance loss of 1\% as acceptable, particularly in regard of a significantly improved anomaly detection performance.} \label{tab:res_segment}
\end{table*}

\subsection{Combining Entropy Maximization and Meta Classification}\label{sec:meta_classification}
Meta classification is the task of discriminating between a true positive prediction and a false positive prediction. For semantic segmentation, this idea was originally proposed in \cite{Rottmann18}. By means of hand-crafted metrics, which are based on dispersion measures, geometry features, or location information, all derived from softmax probabilities, meta classifiers have shown to reliably identify incorrect predictions at segment level. More precisely, connected components of pixels sharing the same class label are considered as segments in this context, and a false positive segment then corresponds to a segment-wise intersection-over-union ($\mathrm{IoU}$) of zero.

The meta classification approach can straightforwardly be adapted to post-process anomaly segmentation masks. This seems particularly reasonable in combination with entropy maximization. Since entropy maximization generally increases the sensitivity towards predicting anomalies, it is possible that the entropy is also increased at pixels belonging to non-anomalous objects. In the latter case, this would yield false positive anomaly instance predictions, which, however, can be identified and discarded afterwards by meta classification. The concept of trading false-positive detection for anomaly detection performance is motivated by \cite{Chan2020}. Moreover, meta classifiers are expected to considerably benefit from entropy maximization, since in the original work \cite{Rottmann18} the entropy as metric has already been observed to be well correlated to the segment-wise $\mathrm{IoU}$.

\begin{figure}[!t]
    \centering
    \captionsetup[subfigure]{labelformat=empty, position=top}
    \subfloat[Without meta classifier]{\includegraphics[width=0.33\textwidth]{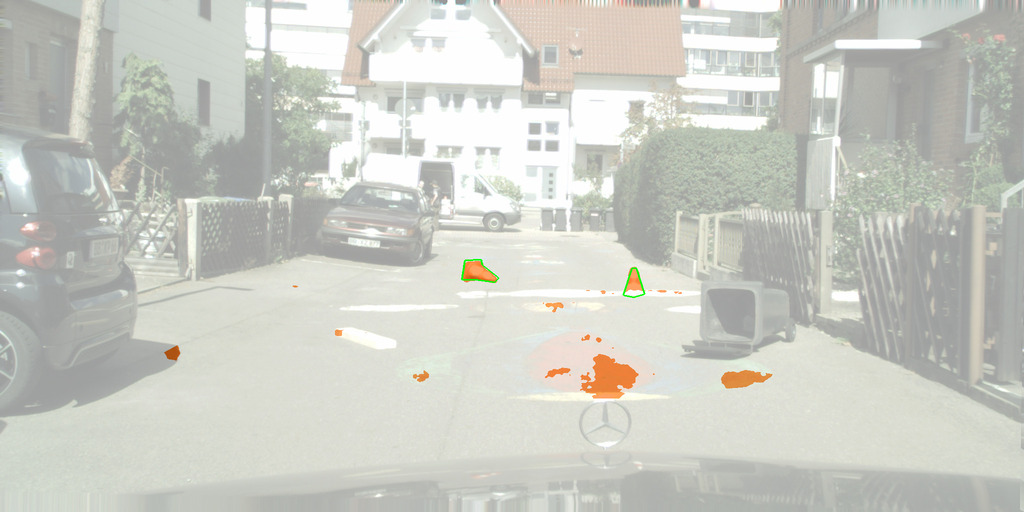}}%
    \subfloat[Prediction quality rating]{\includegraphics[width=0.33\textwidth]{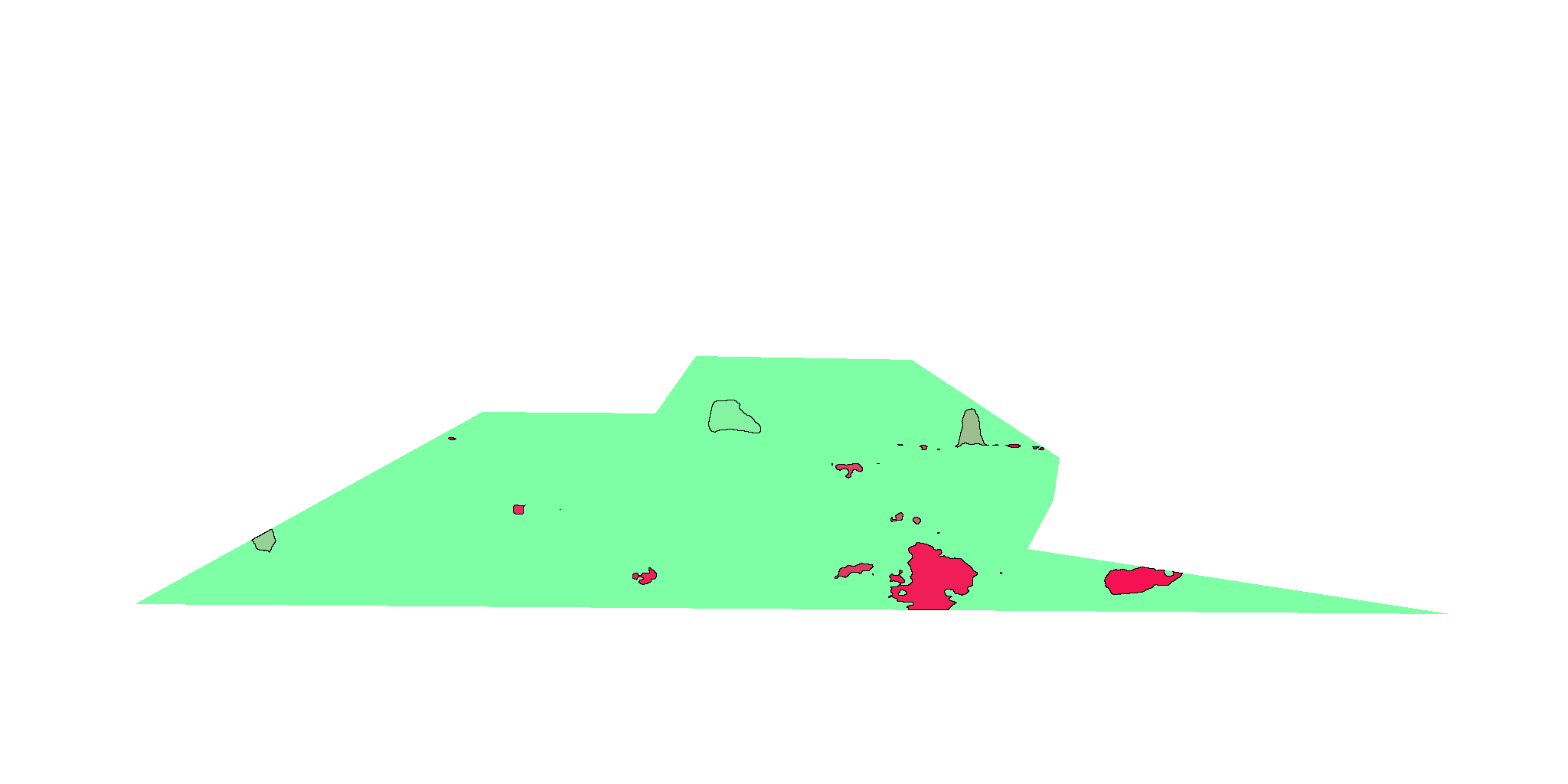}}%
    \subfloat[With meta classifier]{\includegraphics[width=0.33\textwidth]{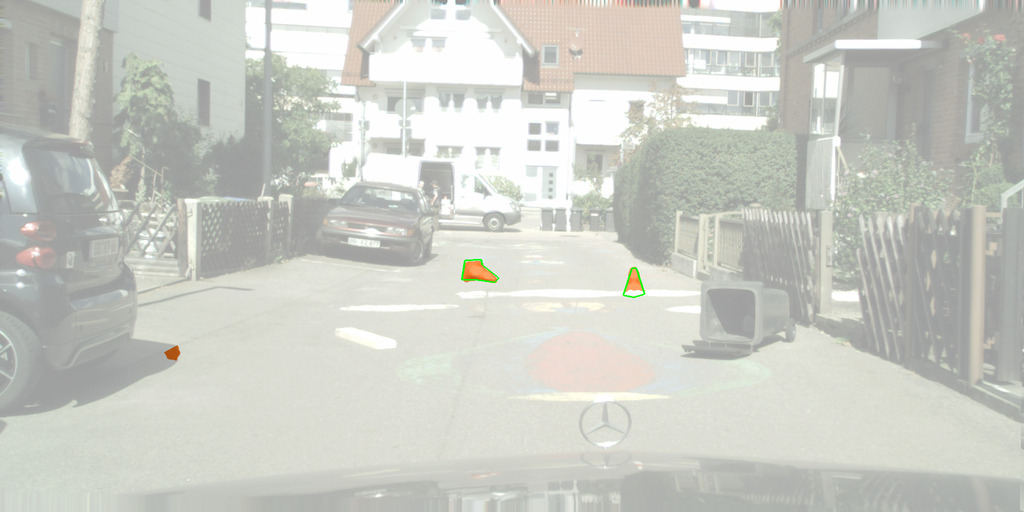}}
    \caption{Meta classification as quality rating of anomaly instance predictions. Before applying the meta classifier (left), the anomaly segmentation mask contains anomaly instance predictions (orange segments), with some false-positives on the road. Based on softmax probabilities, the meta classifier performs a prediction quality rating (middle, red corresponds to poor quality), which is then used to remove false positive anomaly instance predictions (right). Note that the region of interest is restricted to the road, where ground truth anomalous objects (or obstacles) are indicated by green contours.}
    \label{fig:meta_classif}
\end{figure}

In our experiments on LostAndFound \cite{Pinggera2016}, we employ a logistic regression as meta classifier that is applied as a post-processing step on top of softmax probabilities. We observe that the meta classifier is capable of reliably removing false-positive anomaly instance predictions, which in turn significantly improves detection performance of anomalous objects. The meta classification performance is reported in \Cref{tab:res_segment}, a visual example is given in \Cref{fig:meta_classif}.
We note that meta classification is applied to segmentation masks as input. Therefore, the output of the combination of entropy maximization and meta classification does not yield pixel-wise anomaly scores to compare against the methods presented in \Cref{sec:methods}.

The idea of meta classification can even be used to directly identify potential anomalous objects in the semantic segmentation mask, see \cite{Oberdiek20}, which will be subject to discussion in the following section about unsupervised learning of unknown objects.

\section{Discovering and Learning Novel Classes} \label{sec: incremental-learning}

If certain types of anomalies appear frequently, it might be reasonable to include them as additional learnable classes of the segmentation model \cite{Oberdiek20}. In this section, we examine an unsupervised learning method in order to further process anomaly predictions with the goal to produce labels corresponding to novel classes. Afterwards, we will perform an incremental learning approach to train a model on novel classes by means of the retrieved unsupervised labels. This overall procedure has originally been proposed in \cite{uhlemeyer2022towards}.

\subsection{Unsupervised Identification and Segmentation of a Novel Class}
Consider the dataset $ \mathcal{D^\mathrm{test}} \subseteq \mathcal{X} $ of unlabeled images $ \VEC{x} = (x_i)_{i\in\mathcal{I}} \in \mathbb{I}^{H \times W \times 3}$, along with a semantic segmentation network $\VEC{F}: \I^{H \times W \times 3} \to \R^{H \times W \times S} $ trained on the set of classes $\mathcal{S} = \{ 1, \ldots, S \}$. Moreover, let $\VEC{a} = (a_i)_{i\in\mathcal{I}} \in \R^{H \times W}$ denote a score map, as introduced in \Cref{sec:anomaly-segmentation}, which assigns the degree of anomaly to each pixel $i \in \mathcal{I}$ in image $\VEC{x}$. In this section, we employ the unsupervised anomaly segmentation technique described as a three-step procedure in what follows.

\paragraph{1. Image embedding:} Image retrieval methods are commonly applied to construct a database of images that are visually related to a given image. On that account, such methods must quantify visual similarities, \ie measuring the discrepancy or "distance" between images. A simple idea is averaging over the pixel-wise differences. However, this approach is extremely sensitive towards data transformation such as rotation, variation in light, or different resolutions. More advanced approaches make use of visual descriptors that extract the elementary characteristics of the visual contents, \eg color, shape, or texture. These methods are invariant to data transformation, \ie they perform well in identifying images representing the same item. If we want to detect different instances of the same category, deep learning methods represent the state-of-the-art. In this regard, convolutional neural networks (CNNs) achieve very high accuracy in image classification tasks. These networks extract features of the images, that are stable regarding transformations as well as the represented object itself, \ie objects of the same category result in similar feature vectors. We now adapt this idea to identify anomalies that belong to the same class.

Let $\mathcal{K}_{\VEC{a}|\VEC{x}}$ denote the set of connected components within $(a_i^{(\tau)})_{i \in \mathcal{I}}, a_i^{(\tau)} := \VECG{1}_{ \{ a_i \geq \tau \} } ~ \forall~ i\in\mathcal{I}$ for a given threshold $\tau\in \R$, after processing image $\VEC{x}$. Furthermore, let $\mathcal{K} := \bigcup_{x \in \mathcal{X}} \mathcal{K}_{\VEC{a}|\VEC{x}}$ denote
the set of all predicted anomaly components in $\mathcal{D^\mathrm{test}}$. For each component $k\in\mathcal{K}_{\VEC{a}|\VEC{x}}$, we tailor the input $\VEC{x}$ to the image crop $\VEC{x}^{(k)} = (x_i)_{i\in\mathcal{I}'}, \mathcal{I}' \subseteq \mathcal{I}$ by means of the bounding box around $k\in\mathcal{K}_{\VEC{a}|\VEC{x}}$.
By feeding the crop $\VEC{x}^{(k)}$ to an image classification network $\VEC{G}$, we map $\VEC{x}^{(k)}$ onto its feature vector $\VEC{g}^{(k)} := \VEC{G}_{L-1}(\VEC{x}^{(k)}) \in \mathbb{R}^n$, $n\in\mathbb{N}$ 
for all $k\in\mathcal{K}$. Here, $\VEC{G}_{L-1}$ denotes the output of the penultimate layer of $\VEC{G}$.

\paragraph{2. Dimensionality reduction:} Feature vectors extracted by CNNs are usually very high-dimensional. This evokes several problems regarding the clustering of such data. The first issue is known as \emph{curse of dimensionality}, \ie the amount of required data explodes with increasing dimensionality. Furthermore, distance metrics become less precise. Dimensionality reduction approaches project the feature vectors onto a low-dimensional representation, either by feature elimination, selection, or extraction. The latter creates new independent features as a combination of the original vectors and can be further distinguished between linear and non-linear techniques. A linear feature extraction approach, named principal component analysis (PCA) \cite{F.R.S.1901}, aims at decorrelating the components of the vectors by a change of basis, such that they are mostly aligned along the first axes. Thereby, little information is lost if we drop the last components. A more recent non-linear method is t-distributed stochastic neighbor embedding (t-SNE) \cite{JMLR:v9:vandermaaten08a}, which uses conditional probabilities representing pairwise similarities. Let us consider two feature vectors $\VEC{g}^{(k)}$, $\VEC{g}^{(k')}$ with $k,k'\in\mathcal{K}$ and let $\PROBD_{k|k'} \in \I$ denote their similarity under a Gaussian distribution. Employing a Student t-distribution with one degree of freedom in the low-dimensional space then provides a second probability $\mathrm{q}_{k|k'} \in \I$. Hence, t-SNE aims at minimizing the following sum (or
Kullback-Leibler divergence) \cite{JMLR:v9:vandermaaten08a}
\begin{equation}
    \sum_{k\in\mathcal{K}}\sum_{k'\in\mathcal{K}}\PROBD_{k|k'} \log \left( \frac{\PROBD_{k|k'}}{\mathrm{q}_{k|k'}} \right)
\end{equation}
using gradient descent. We first perform dimensionality reduction via PCA, which is then followed by t-SNE.
In our experiments, we observe that this combination of methods improves the effectiveness of mapping anomaly predictions onto a two-dimensional embedding space. Here, the embedding ideally creates neighborhoods of visually related anomalies.

\paragraph{3. Novelty segmentation:}  If anomalies of the same category are detected more frequently, they are expected to form a bigger cluster in the embedding space. Those clusters can be identified by employing algorithms such as density-based spatial clustering of applications with noise (DBSCAN) \cite{Ester1996}. 
This algorithm supports the idea of non-supervision since it does not require any information of the potential anomaly data, such as \eg the number of clusters. To this end, DBSCAN divides data points into core points, border points, and noise, depending on the size of the neighborhood $\varepsilon\in[0,\infty)$ and the minimal number of a core point's neighbors $\delta\in\mathbb{N}$. 

More precisely, let $\tilde{\VEC{g}}^{(k)} \in \R^2$ denote the two-dimensional representation of $\VEC{x}^{(k)}$. Then, $\tilde{\VEC{g}}^{(k)}$ is considered as a core point, if the corresponding point-wise density $\rho(\tilde{\VEC{g}}^{(k)}) := |\{ \tilde{\VEC{g}}^{(k')} ~:~\Vert \tilde{\VEC{g}}^{(k)} - \tilde{\VEC{g}}^{(k')}\Vert < \varepsilon, k'\in\mathcal{K}\}| \geq \delta$, \ie the $\varepsilon$-neighborhood of $\tilde{\VEC{g}}^{(k)}$ contains at least $\delta$ points including itself.
We denote the neighborhood of a core point by $\tilde{\VEC{g}}^{(\mathring{k})}$, which corresponds to a component $\mathring{k} \in \mathcal{K}$, as $ B_{\mathring{k}} := \{ \tilde{\VEC{g}}^{(k')} :  \Vert \tilde{\VEC{g}}^{(\mathring{k})} - \tilde{\VEC{g}}^{(k')}\Vert < \varepsilon, k'\in\mathcal{K} \}.$ If $\tilde{\VEC{g}}^{(k)}$ is not a core point but belongs to a core point's neighborhood, we call it a border point. Otherwise, \ie if $\tilde{\VEC{g}}^{(k)}$ is neither a core point nor within a core point's neighborhood, we call it noise. 

Finally, a cluster $\mathcal{C}_j \subset \mathcal{K}, j \in \mathcal{J} := \{ 1, \ldots, J \}$ of components is formed by merging overlapping neighborhoods $B_{\mathring{k}}$, yielding $J\in\N$ clusters in total. In other words, clusters
are formed from connected core points and their neighborhoods' border points.
Given $\rho(\tilde{\VEC{g}}^{(k)})$, we can determine the cluster density of 
$\mathcal{C}_j$, \eg $$ \text{as the maximum~~} \max_{k\in\mathcal{C}_j} \rho(\tilde{\VEC{g}}^{(k)})\;, \text{~~or as the average~~} \frac{1}{|\mathcal{C}_j|}\sum_{k\in\mathcal{C}_j} \rho(\tilde{\VEC{g}}^{(k)})\;.$$
The cluster $\mathcal{C}^*\subset \mathcal{K}$, which is the cluster of highest density given a sufficient cluster size, is then selected to be further processed.  
To this end, let us consider the predicted segmentation mask $\VEC{F}(\VEC{x})=\VEC{m} = (m_i)_{i\in\mathcal{I}}$, where $m_i = \arg\max_{s\in\mathcal{S}} y_{i,s}, i \in \mathcal{I}$. The pseudo labels $\tilde{\VEC{y}} = (\tilde{y}_i)_{i\in\mathcal{I}}$ for the originally unlabeled $\VEC{x}$ are then obtained by setting $\tilde{y}_i = S + 1$ if pixel location $i$ belongs to a component $k \in \mathcal{C}^*$, and $\tilde{y}_i = m_i$ otherwise.

\subsection{Class-Incremental Learning} \label{sec:incremental-learning-des}

Let $\tilde{\mathcal{Y}}$ denote the set of pseudo labels, then the training data for some novel class $S+1$ can be represented by $\mathcal{D}^{S+1} \subseteq \mathcal{D}^\mathrm{novel} \times \tilde{\mathcal{Y}} $, where $\mathcal{D}^\mathrm{novel}$ denotes the set of previously-unseen images containing novel classes. By extending the semantic segmentation network $\VEC{F}$ to $\VEC{F}^+: \I^{H \times W \times 3} \to \R^{H \times W \times (S+1)}$ and retraining $\VEC{F}^+$ on $\mathcal{D}^{S+1}$, we perform incremental learning to add a novel and previously unknown class to the semantic space of $\VEC{F}$.

\paragraph{Regularization:} Knowledge distillation is a subcategory of regularization strategies aiming to mitigate a catastrophic forgetting, \ie these strategies try to mitigate performance loss on the previously-learned classes $\mathcal{S} = \{1, \ldots, S\}$ while learning the additional class $S+1$. In \cite{Michieli2019}, the authors adapted incremental learning techniques to the task of semantic segmentation. Among others, they introduced the overall objective 
\begin{equation}
    \label{eq:combined_loss}
    J^\mathrm{total}(\VEC{x},\tilde{\VEC{y}}) = (1-\lambda)\, J^\mathrm{CE}(\VEC{F}^+(\VEC{x}), \tilde{\VEC{y}}) + \lambda\, J^\mathrm{D}(\VEC{F}^+(\VEC{x}), \VEC{F}(\VEC{x})), ~~ \lambda \in \I \; ,
\end{equation}
where $(\VEC{x},\tilde{\VEC{y}}) \in \mathcal{D}^{S+1}$. Here, $J^\mathrm{CE}$ denotes the common cross-entropy loss over the enlarged set of class indices $\mathcal{S}^+ := \{1,\ldots,S+1\}$ and $J^\mathrm{D}$ the distillation loss. The latter loss is defined as 
\begin{equation}
    J^\mathrm{D}(\VEC{F}^+(\VEC{x}), \VEC{F}(\VEC{x})) := -\frac{1}{H\cdot W} \sum_{i\in\mathcal{I}}\sum_{s\in\mathcal{S}} \mathrm{softmax}_s(\VEC{y}_{i})\log(\mathrm{softmax}_s(\VEC{y}^+_{i}))
\end{equation}
with $\VEC{y} = \VEC{F}(\VEC{x})$ and $\VEC{y}^+ = \VEC{F}^+(\VEC{x})$. 
Knowledge distillation can be further improved by freezing the weights of the encoder part of $\VEC{F}^+$ during the training procedure \cite{Michieli2019}. 

\begin{figure}[!t]
    \captionsetup[subfigure]{labelformat=empty}
    \centering
    \subfloat[Segmentation prediction of the initial model]{\includegraphics[width=.48\textwidth]{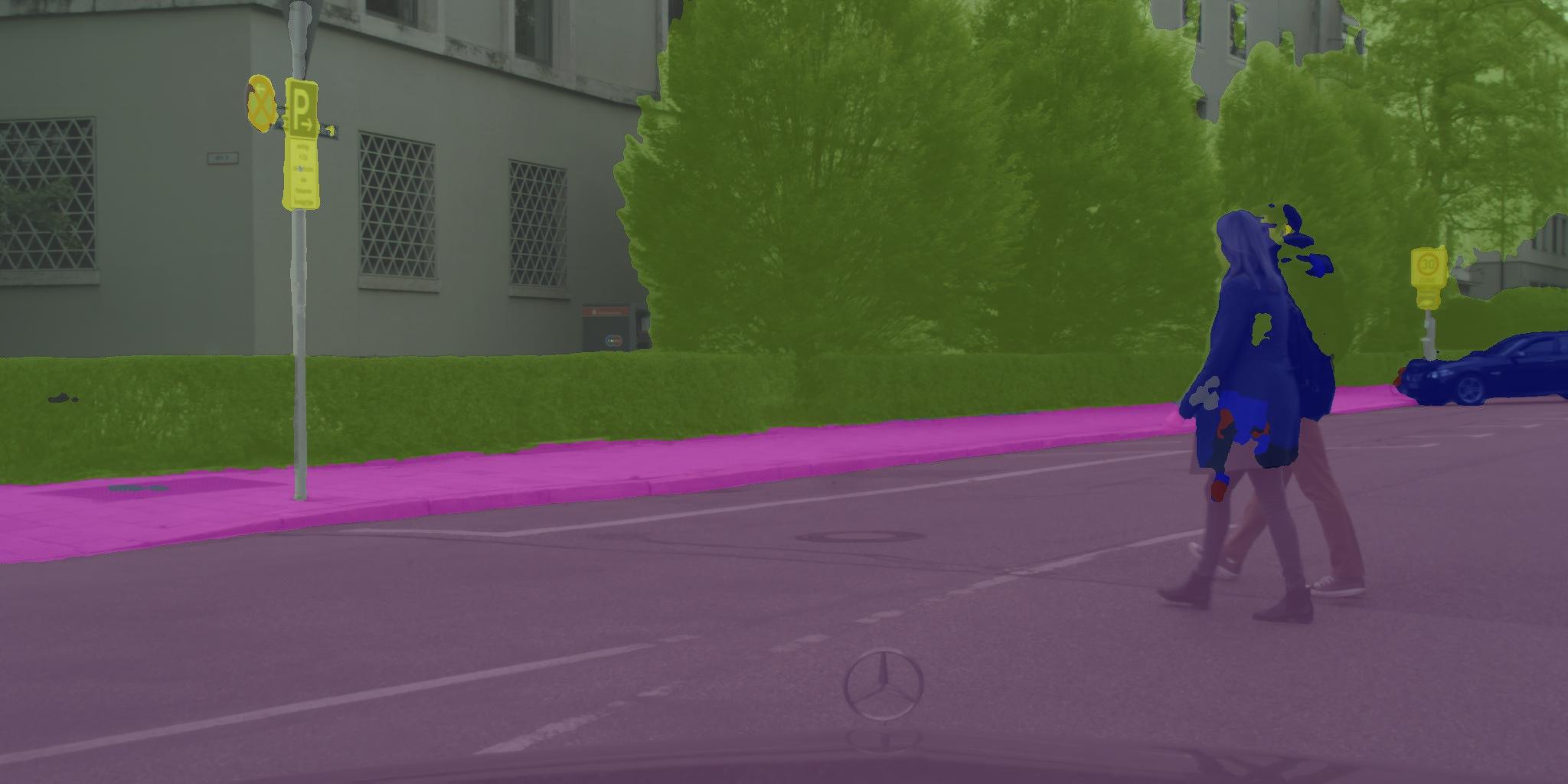}}~
    \subfloat[Prediction quality rating]{\includegraphics[width=0.48\textwidth]{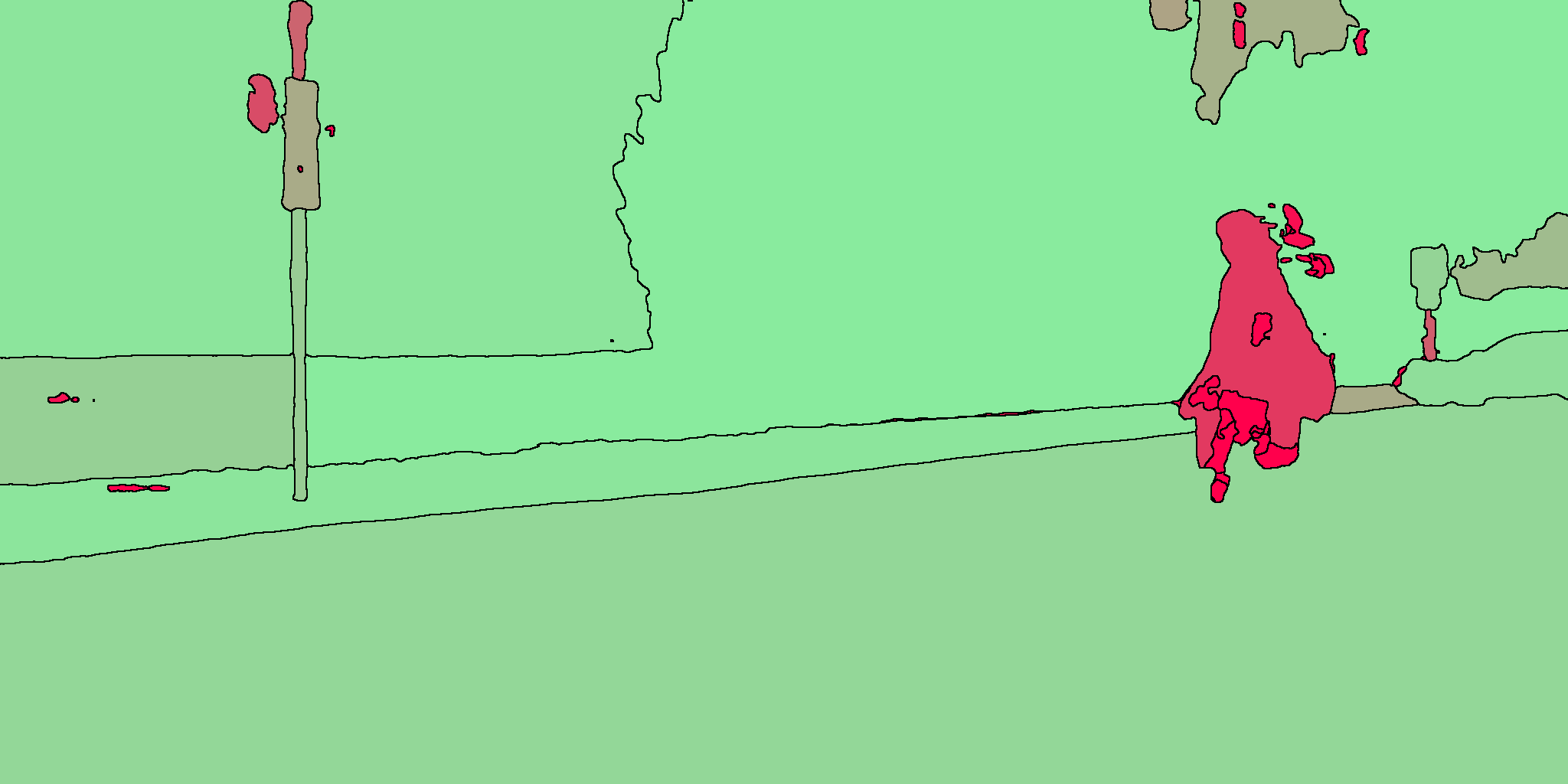}}
    \caption{Predicted semantic segmentation mask of the initial model's prediction (left) and corresponding segment-wise quality estimation (right) for one example from the Cityscapes test split. Green color indicates a high segment-wise $\mathrm{IoU}$, red color indicates a low one.}
    \label{fig:human_ms}
\end{figure}

\paragraph{Rehearsal:} If the original training data $\mathcal{D}^\mathrm{train} \subseteq \mathcal{X} \times \mathcal{Y}$ of network $\VEC{F}$ is available, in incremental learning such data is usually
re-integrated into the training set of the extended network $\VEC{F}^+$, \ie the training samples are drawn from $\mathcal{D}^\mathrm{train} \cup \mathcal{D}^{S+1}$. To save computational costs of training and to balance the amount of old and new training data, established methods, \eg \cite{Robins1995}, only use a subset of $\mathcal{D}^\mathrm{train}$. 
This subset is typically obtained by randomly sampling a set from $\mathcal{D}^\mathrm{train}$ that matches the size of $|\mathcal{D}^{S+1}|$.

In combination with knowledge distillation, rehearsal strategies can be employed to mitigate a loss of performance on classes that are related to the novel class. This issue may arise \eg through visual similarity such as between classes like \emph{bus} and \emph{train}, or due to class affiliation as in the case of \emph{bicycle} and \emph{rider}.
Relevant classes can be identified by their frequency of being predicted on the relabeled pixels, \ie
\begin{equation}
    \nu_s^\mathrm{tot} := \sum_{(\VEC{x},\tilde{\VEC{y}})\in\mathcal{D}^{S+1}} \big\vert \{i\in\mathcal{I}~|~m_i = s~\wedge~\tilde{y}_i= S+1\} \big\vert ~~ \forall ~s\in\mathcal{S} \; ,
\end{equation}
and hence
\begin{equation}
    \nu_s^\mathrm{rel} := \frac{\nu_s^\mathrm{tot}}{\sum_{s'\in\mathcal{S}}\nu_{s'}^\mathrm{tot}}~~ \forall ~ s\in\mathcal{S} \; .
\end{equation}
The subset of $\mathcal{D}^\mathrm{train}$ is then randomly sampled under the constraint that there are at least $\nu_s^\mathrm{rel}|\mathcal{D}^{S+1}|$ images containing the class $s$ for all $s\in\mathcal{S}$.

\subsection{Experiments and Evaluation}

\begin{figure}[!t]
    \captionsetup[subfigure]{labelformat=empty}
    \centering
    \subfloat{\includegraphics[width=0.99\textwidth]{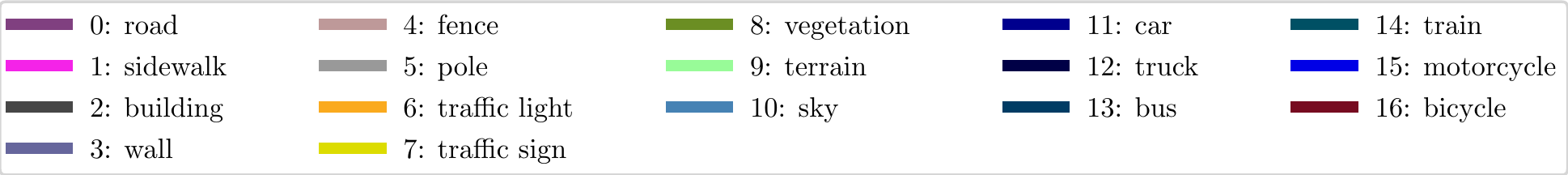}}\\
    \subfloat{\includegraphics[width=0.70\textwidth]{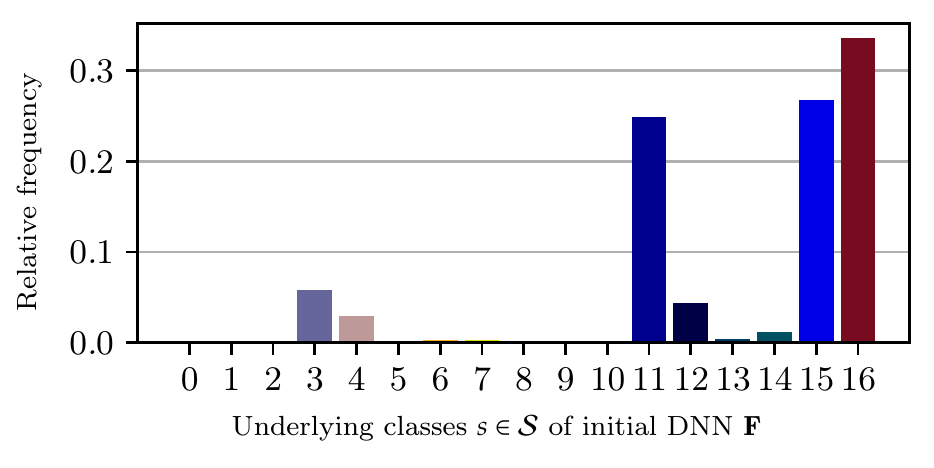}}
    \caption{Relative frequency of old classes being predicted by the initial model on pixels that are assigned to the novel class. Thus, the subset of $\mathcal{D}_\mathrm{CS}^\mathrm{train}$ included in the retraining should mainly involve bicycles, motorcycles, and cars.}
    \label{fig:preds_human}
\end{figure}

In the following experiments, we will employ a \texttt{DeepLabV3+} \cite{10.1007/978-3-030-01234-2_49} model with an underlying \texttt{WiderResNet38} \cite{Zhu2019} backbone for semantic segmentation. This network is initially trained on a set of $17$ classes, which we will extend by a novel class. The already trained classes are the Cityscapes training classes except \emph{pedestrian} and \emph{rider}, \ie we exclude any \emph{human} in the training process of our initial semantic segmentation network $\VEC{F}$.

The initial model was trained on the Cityscapes \cite{Cordts2016} training data. For the incremental learning process, we use a portion of those data and combine them with the generated disjoint training set $\mathcal{D}^{S+1}$ containing previously unseen images and pseudo labels on novel objects. Here, the images from $\mathcal{D}^{S+1}$ are drawn from the Cityscapes test data. 
For evaluation purposes, we use the Cityscapes validation data.
Hence, during the incremental learning process only known objects are presented to the model except humans and a few instances, such as the ego-car or mountains in an image background, belonging to the Cityscapes void category.

We use the idea of meta classification, similarly as introduced in \Cref{sec:meta_classification}, to rate the prediction quality of predicted semantic segmentation masks. 
Here, the meta task is used to estimate the segment-wise IoU first, see \Cref{fig:human_ms}, on which we apply thresholding (at $\tau=0.5$) to determine potential anomalies, \cf \cite{Oberdiek20}.
We employ gradient boosting as meta model, which achieves a coefficient of determination of $\mathrm{R}^2 = 82.51\%$ in estimating the segment-wise IoU on the Cityscapes validation split.

\begin{figure}[!t]
    \captionsetup[subfigure]{labelformat=empty}
    \centering
    \subfloat[Prediction of initial model]{\includegraphics[width=0.31\textwidth]{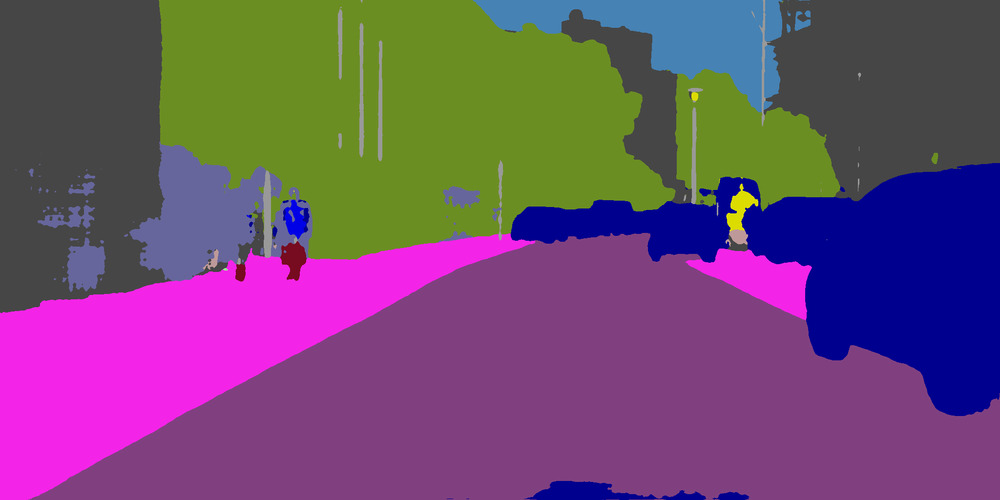}}~
    \subfloat[Novel class in validation image]{\includegraphics[width=0.31\textwidth]{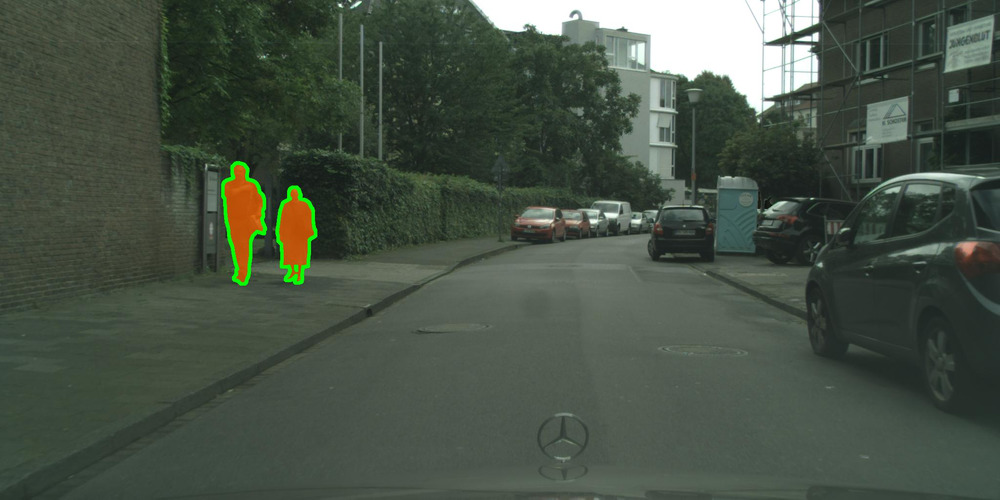}}~
    \subfloat[Prediction of extended model]{\includegraphics[width=0.31\textwidth]{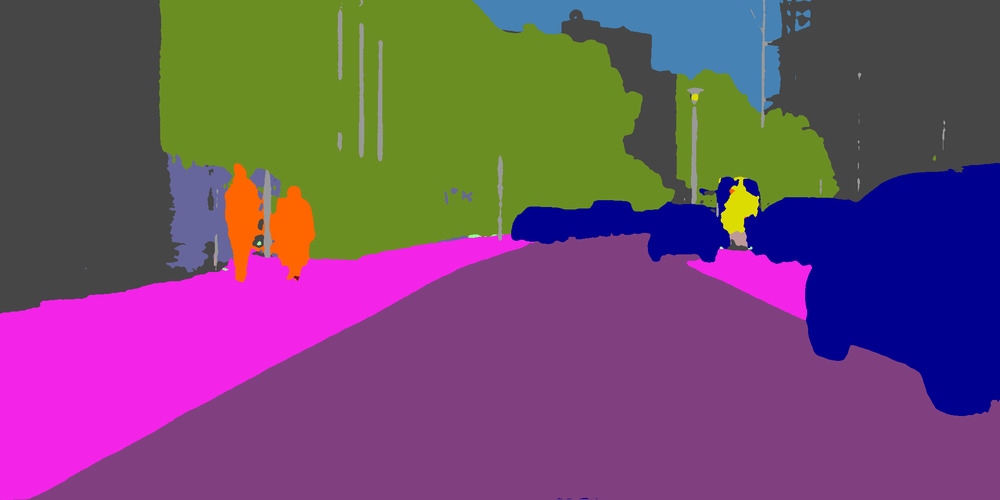}}
    \caption{Comparison of the predicted semantic segmentation masks before (left) and after (right) adapting the model to the novel \emph{human} class (orange) for one example of the Cityscapes validation split (middle). Here, the novel components are highlighted in orange, green contours indicate the ground truth annotation of the novelty.}
    \label{fig:human}
\end{figure}

\newcolumntype{g}{>{\columncolor{gray!50}}c}
\begin{table}[t]
    \centering
    \resizebox{1\textwidth}{!}{
    \tabcolsep1.75pt
    \begin{tabular}{cl||c|c|c|c|c|c|c|c|c|c|c|g|c|c|c|c|c|c||c|c} 
        \hline
         & metric & \rotatebox[origin=l]{90}{road} & \rotatebox[origin=l]{90}{sidewalk} & \rotatebox[origin=l]{90}{building}& \rotatebox[origin=l]{90}{wall}& \rotatebox[origin=l]{90}{fence}& \rotatebox[origin=l]{90}{pole}& \rotatebox[origin=l]{90}{traffic light}& \rotatebox[origin=l]{90}{traffic sign}& \rotatebox[origin=l]{90}{vegetation}& \rotatebox[origin=l]{90}{terrain}& \rotatebox[origin=l]{90}{sky}& \rotatebox[origin=l]{90}{human}& \rotatebox[origin=l]{90}{car}& \rotatebox[origin=l]{90}{truck}& \rotatebox[origin=l]{90}{bus}& \rotatebox[origin=l]{90}{train}& \rotatebox[origin=l]{90}{motorcycle}& \rotatebox[origin=l]{90}{bicycle}& \rotatebox[origin=l]{90}{mean excl. human~~}& \rotatebox[origin=l]{90}{mean incl. human~}  \\\hline\hline
         \multirow{3}{*}{\rotatebox[origin=c]{90}{before}} & $\mathrm{IoU}$ & 97.34 & 80.63 & 88.91 & 47.24 & 51.03 & 52.90 & 55.44 & 66.66 & 89.95 & 56.29 & 93.76 & 00.00 & 90.61 & 69.66 & 76.90 & 70.35 & 24.45 & 54.57 & 68.63 & 64.82\\
         & $\PRECISION$ & 98.35 & 89.39 & 92.80 & 74.57 & 66.76 & 72.68 & 75.04 & 86.22 & 93.60 & 77.66 & 96.38 & 00.00 & 92.97 & 80.23 & 88.59 & 83.33 & 28.57 & 59.30 & 79.79 & 75.36\\
         & $\RECALL$ & 98.96 & 89.16 & 95.50 & 56.32 & 68.41 & 66.02 & 67.98 & 74.61 & 95.85 & 67.17 & 97.18 & 00.00 & 97.27 & 84.09 & 85.35 & 81.87 & 62.92 & 87.24 & 80.94 & 76.44\\\hline\hline
         \multirow{3}{*}{\rotatebox[origin=c]{90}{after}} & $\mathrm{IoU}$ & 97.46 & 80.78 & 89.30 & 47.48 & 49.31 & 53.25 & 55.28 & 65.72 & 90.17 & 54.53 & 93.47 & 41.42 & 91.21 & 69.30 & 72.52 & 62.06 & 30.45 & 57.72 & 68.24 & 66.75\\
         & $\PRECISION$ & 98.68 & 89.31 & 93.11 & 78.33 & 69.48 & 73.74 & 76.02 & 88.99 & 94.21 & 75.66 & 95.69 & 59.73 & 95.26 & 84.88 & 87.26 & 91.68 & 64.38 & 76.01 & 84.28 & 82.91\\
         & $\RECALL$ & 98.75 & 89.43 & 95.62 & 54.67 & 62.95 & 65.70 & 66.96 & 71.54 & 95.46 & 66.13 & 97.57 & 57.48 & 95.45 & 79.06 & 81.11 & 65.76 & 36.61 & 70.57 & 76.08 & 75.05\\\hline
    \end{tabular}}
    \caption{Evaluation of the Cityscapes validation split before and after incremental learning the novelty \emph{human} (highlighted in gray) with knowledge distillation and rehearsal. The classes \emph{pedestrian} and \emph{rider} are aggregated to the novel class \emph{human}. All other classes are treated as background, \ie they are ignored during training, regarding the data from $\mathcal{D}^{S+1}$.}
    \label{tab:human_eval}
\end{table}

In accordance to \Cref{sec: outlier-statistics} and as already observed in \Cref{sec:meta_classification}, the softmax entropy is again one of the main metrics included in the meta model to identify anomalous predictions. Thus, the entropy shows to have great impact on meta classification performance, which, similarly, has also been observed in \cite{Chan2020, Chan2020entropy}.

Given anomaly segmentation masks, we perform image embedding using the encoder of the image classification network \texttt{DenseNet201} \cite{Huang_2017}, that is pretrained on ImageNet \cite{DBLP:conf/cvpr/DengDSLL009}. Next, we reduce the dimensionality of the resulting feature vectors to $50$ via PCA and further to $2$ by applying t-SNE. In \cite{Oberdiek20}, a qualitative and quantitative evaluation of different embedding approaches is provided. Note that t-SNE is non-deterministic, \ie we obtain slightly different embedding spaces for different runs. In our experiments, employing DBSCAN with parameters $\varepsilon=2.5$ and $\delta=15$ produces a \emph{human}-cluster including $91$ components from $76$ different images. The most frequently predicted class of these components are \emph{car}, \emph{motorcycle}, and \emph{bicycle} with $\nu_{11}^\mathrm{rel} = 24.84\%$, $\nu_{15}^\mathrm{rel} = 26.69\%$ and $\nu_{16}^\mathrm{rel} = 33.53\%$, respectively, see \Cref{fig:preds_human}.

We train the extended model $\VEC{F}^+$ as described in \Cref{sec:incremental-learning-des} for $70$ epochs, weighting the loss functions in \Cref{eq:combined_loss} equally, \ie $\lambda=0.5$. The extended model is capable of retaining its initial knowledge by achieving an $\mIoU$ score of $68.24\%$ on the old classes when evaluating on the Cityscapes validation data. This yields a marginal loss of only $0.39\%$ compared to the initial model $\VEC{F}$. At the same time, $\VEC{F}^+$ predicts the novel human class with a class $\mathrm{IoU}$ of $41.42\%$, without a single annotated human instance in the training data $\mathcal{D}^{S+1}$. A visual example of the applied unsupervised novelty segmentation approach is provided in \Cref{fig:human}, more details on the numerical evaluation is given in \Cref{tab:human_eval}.

\subsection{Outlook on Improving Unsupervised Learning of Novel Classes}

\begin{figure}[!t]
    \captionsetup[subfigure]{labelformat=empty} 
    \centering
    \subfloat[]{\includegraphics[width=0.49\textwidth]{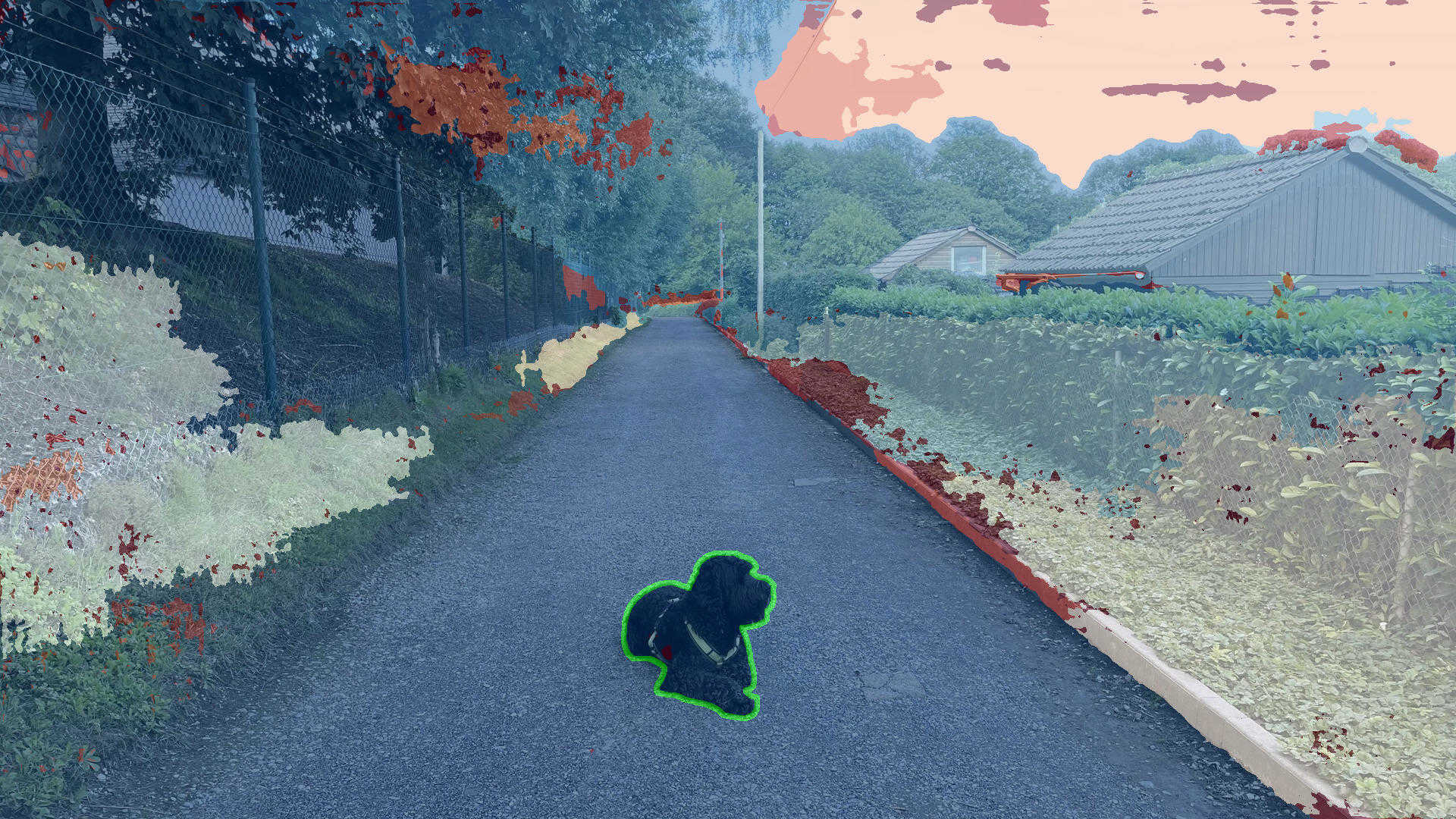}}~
    \subfloat[]{\includegraphics[width=0.49\textwidth]{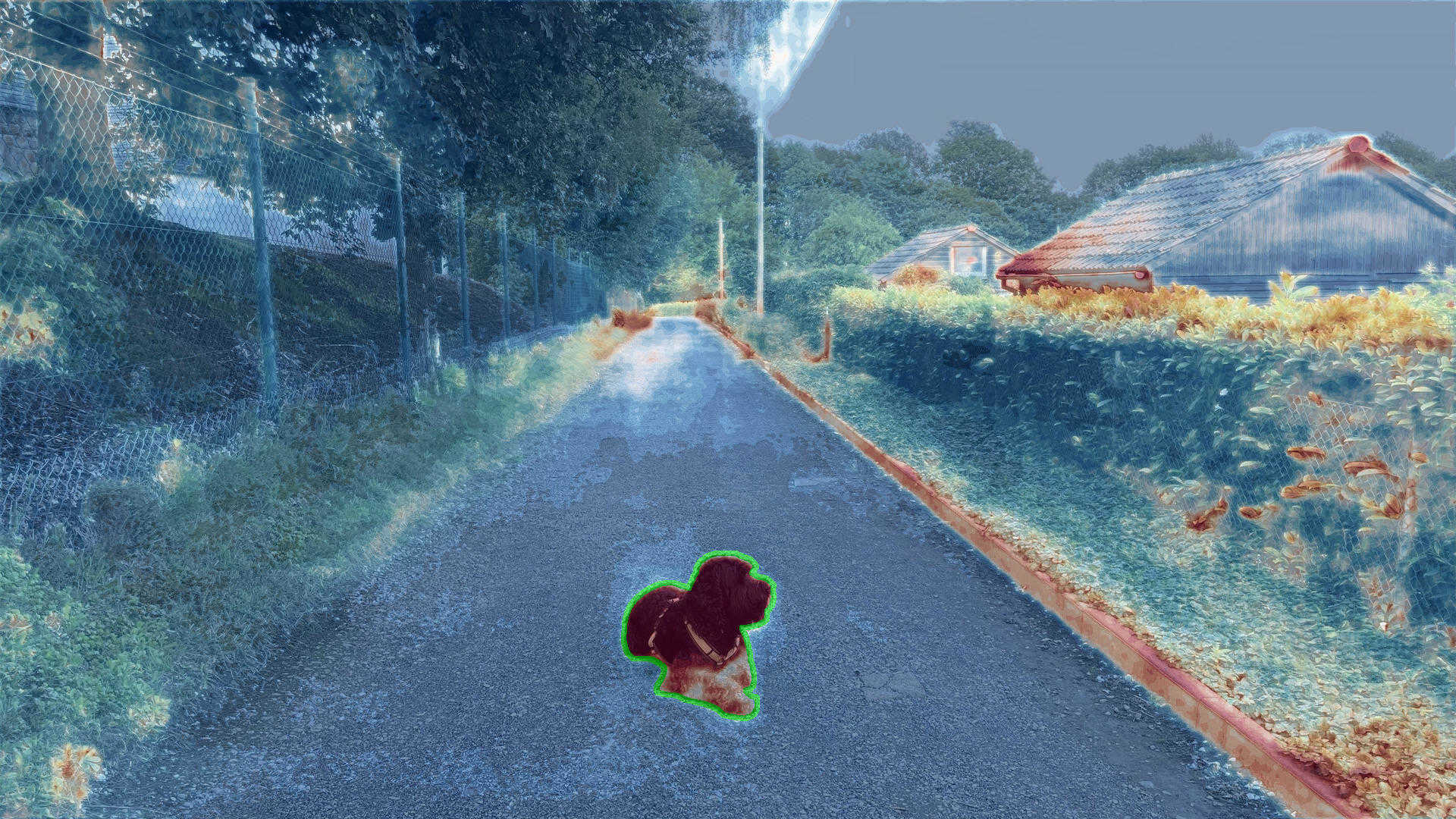}}
    \caption{A comparison of anomaly scores obtained by meta classification (left) and entropy (right) on an image from RoadObstacle21. The dog on the image is the anomaly of interest (indicated by green contours), which would have been overlooked by meta classification but entirely detected by the entropy.}
    \label{fig:entropy-vs-metaseg}
\end{figure}

In the preliminary experiments presented in this section, we have demonstrated that a semantic segmentation network can be extended by a novel class in an unsupervised fashion. As a basis to start, the investigated unsupervised learning approach requires anomaly segmentation masks. Currently, these are obtained by meta classification \cite{Rottmann18}, which is, however, not a method specifically tailored for the task of anomaly segmentation. In other words, the obtained masks are possibly inaccurate.
To be even more precise on the limitation of plain meta classification, this method is only able to find anomalies when the segmentation model produces a (false positive) object prediction on those anomalies. By design, meta classifiers cannot find overlooked instances, \eg obstacles on the road that also have been classified as road. As an illustration of this issue, we refer to \Cref{fig:entropy-vs-metaseg}.

Having now several anomaly segmentation methods at hand, which we, \eg introduced in \Cref{sec:methods}, it seems obvious to replace the underlying anomaly segmentation method by a more sophisticated one as future work. In this light, given the decent performance of the examined unsupervised learning method relying only on meta classification, combining this method with entropy maximization represents a particularly promising approach to further improve the presented novelty training, \cf again \Cref{sec:meta_classification}.

\section*{Conclusions}

Semantic segmentation as a supervised learning task is typically performed by models that operate on a given set containing a fixed number of classes. This is in clear contrast to the open world scenarios to which practitioners contemplate the usage of segmentation models. There are important capabilities that standard segmentation models do not exhibit. Among them is the capability to know when they face an object of a class they have not learned -- \ie to perform anomaly segmentation -- as well as the capability to realize that similar objects, presumably of the same (yet unknown) class, appear frequently and should be learned either as a new class or be attributed to an existing one. In this work, we have seen first promising results for two tasks, for anomaly segmentation as well as for the detection and unsupervised learning of new classes.

For anomaly segmentation, we considered a number of generic baseline methods stemming from image classification as well as some recent anomaly segmentation methods. Since the latter clearly outperforms the former, this stresses the need for the development of methods specifically designed for anomaly segmentation. With the entropy maximization method, we have demonstrated empirically as well as theoretically that good proxies in combination with training on anomaly examples for high entropy could be key to anomaly segmentation capabilities.
Particularly on the challenging RoadObstacle21 dataset with diverse street scenarios, entropy maximization yields great performance, outperforming many other established methods. 
While there exists a moderate number of datasets for anomaly segmentation, there is clearly still the need of additional datasets.
The number of possible unknown object classes not covered by these datasets is evidently enormous. Furthermore, also the vast variety of possible environmental conditions and further domain shifts that may occur, possibly also in combination with unknown objects, continuously demand their exploration.

For detection and unsupervised learning of new classes, we have demonstrated in preliminary experiments that a combination of well-established dimensionality reduction and clustering methods along with the advanced uncertainty quantification method for semantic segmentation called MetaSeg is well able to detect unknown classes of which objects appear relatively frequently in a given test set. Indeed, MetaSeg can also be used to define segmentation proposals for pseudo ground-truths of new classes, which can also be learned incrementally by the segmentation model. For the considered scenario of subsequently learning humans within the Cityscapes dataset, this approach yields an reasonable segmentation performance on the novel class without significantly losing performance on the original classes. This examined methodology may help to incorporate new classes into existing models with low human labeling effort. The necessity for this will occur repeatedly in future. An example are the electric scooters that have recently arisen in several metropolitan areas across the globe. This is an example for a global phenomenon. However, also local phenomena, such as boat trailers at the coast, could be of interest. Such classes could be initially incorporated into an existing model using the presented methodology in this work. Afterwards, the initial performance could be further improved with active learning approaches, such as presented in \cite{colling2020metabox}, still requiring only a small amount of human labeling effort. It is also an open question, to which extent the proposed method can be used iteratively to improve the performance on a new class. Also for this track of research, the lack of data for pursuing that task is a limiting factor as of now.

\subsection*{Acknowledgments} 
\begin{wrapfigure}[6]{r}{0.4\textwidth}
  \begin{center}
    \vspace{-40pt}
    \href{https://www.bmwi.de/Navigation/EN/Home/home.html}{\includegraphics[width=0.25\textwidth]{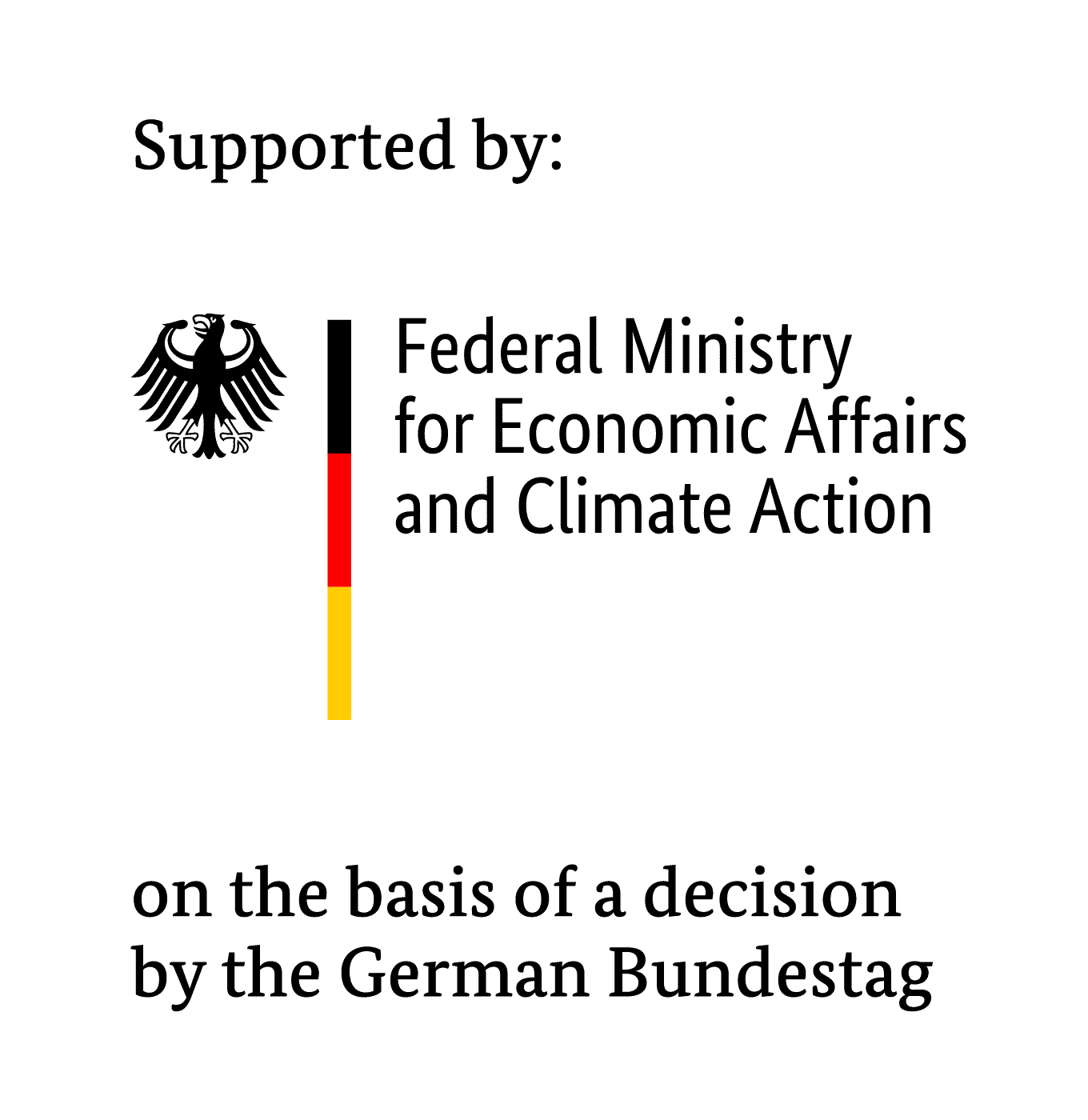}}
  \end{center}
\end{wrapfigure}
The research leading to these results is funded by the German Federal Ministry for Economic Affairs and Climate Action within the projects "\href{https://www.ki-absicherung-projekt.de/en/}{KI Absicherung - Safe AI for Automated Driving}", grant no.\ 19A19005R, and "\href{https://www.ki-deltalearning.de/en/}{KI Delta Learning - Scalable AI for Automated Driving}", grant no.\ 19A19013Q, respectively.
The authors would like to thank the consortia for the successful cooperation.

The authors also acknowledge the Gauss Centre for Supercomputing e.V.\ (\href{https://www.gauss-centre.eu}{www.gauss-centre.eu}) for funding this project by providing computing time through the John von Neumann Institute for Computing (NIC) on the GCS Supercomputer JUWELS at Jülich Supercomputing Centre (JSC).